\RequirePackage{fix-cm}
\documentclass[smallcondensed]{svjour3}     
\usepackage{graphicx}
\usepackage{amsmath,amsfonts,amssymb, float, bbm} 
\usepackage{multicol, multirow, subfigure, lscape}
\usepackage{csquotes}
\usepackage[]{algorithm2e} 
\usepackage[textwidth=1.4in,shadow,colorinlistoftodos]{todonotes}

\newcommand{\xtil}{\tilde{\pmb{x}}}
\newcommand{\rBrackets}[1]{\left( #1 \right)}

\usepackage{listings}
\begin{document}

\title{Detection of Interacting Variables for Generalized Linear Models via Neural Networks  
}

\titlerunning{Interaction detection for GLMs via neural networks}        

\author{Yevhen Havrylenko \and
        Julia Heger
}


\institute{Yevhen Havrylenko (corresponding author)\at
              University of Copenhagen, Department of Mathematical Sciences \\
              Universitetsparken 5, 2100 Copenhagen, Denmark\\
              \email{yh@math.ku.dk} \\
              ORCID: 0000-0002-1877-6072
           \and
           Julia Heger \at
              University of Augsburg, Chair of Analytics \& Optimization\\
              Universit\"atsstra\ss{}e 2, 86159 Augsburg, Germany\\
              \email{julia.heger@uni-a.de}
}

\date{Received: date / Accepted: date}

\maketitle

\begin{abstract}
The quality of generalized linear models (GLMs), frequently used by insurance companies, depends on the choice of interacting variables. The search for interactions is time-consuming, especially for data sets with a large number of variables, depends much on expert judgement of actuaries, and often relies on visual performance indicators. Therefore, we present an approach to automating the process of finding interactions that should be added to GLMs to improve their predictive power. Our approach relies on neural networks and a model-specific interaction detection method, which is computationally faster than the traditionally used methods like Friedman H-Statistic or SHAP values. In numerical studies, we provide the results of our approach on artificially generated data as well as open-source data.

\end{abstract}

\section{Introduction} \label{intro}
Insurance companies usually apply generalized linear models (GLMs) to predict insurance claim counts due to the interpretability of these models.  GLMs are constantly improved by pricing actuaries via sophisticated choice of variable interactions. This process is time-consuming, depends much on expert judgement, and relies on visual performance indicators. These aspects motivate the usage of machine learning (ML) techniques for improving the performance of GLMs by finding the next-best interaction to be added to the GLM. Such automation of the manual and mainly visual process of fine-tuning GLMs could save much time for actuaries, especially in the case of big data sets with dozens of variables, e.g., in motor third-party liability (MTPL) insurance.

In this paper, we propose a methodology for the detection of the next-best interaction that is missing in a benchmark GLM. {We aim at improving an arbitrary but fixed existing benchmark GLM instead of creating a new GLM from scratch.} Building a new GLM may necessitate drastic changes in the tariff of the MTPL insurance. Large changes in tariffs are not desired by insurance companies for their existing business lines. Instead, GLMs need to be improved gradually.

The approach we suggest has three steps. First, a combined actuarial neural network (CANN) is trained. This model is introduced in the actuarial context in \cite{wuethrich2019cann} and can be seen as a combination of a (linear) benchmark-GLM predictor and a (non-linear) neutral network predictor into a single neural network using a skip connection for the GLM. Second, the strength of each pairwise interaction learned by the CANN model is quantified and the interactions are ranked by their strength using a neural interaction detection (NID) algorithm. This algorithm is introduced in \cite{tsang2017detecting} for fully connected feed-forward neural networks and adjusted by us to CANN models. Third, the top-ranked interactions are analyzed with the help of mini-GLMs and the next-best interaction to be included in the benchmark GLM is identified.

We show the performance of our approach on an artificially created data set, where the true interactions are known from the data-generation mechanism, and on an open-source French MTPL data set, which has been analyzed in many academic sources, e.g., \cite{schelldorfer2019}, \cite{noll2018case}, \cite{ferrario2018insights}, \cite{henckaerts2022}, \cite{wuethrich2022}, etc. Finally, we comment on the advantages of our methodology for big MTPL data sets with millions observations and dozens of features, which are common for larger insurance companies.

\textbf{Literature overview}. GLMs are introduced by \cite{Nelder1972} as a generalization of a linear regression model with a normally distributed response variable. Since then, GLMs became an important and popular tool in insurance pricing, especially MTPL insurance. For more information, see  \cite{denuit2007}, \cite{Ohlsson2010}, \cite{wuethrich2022}.

The process of finding the best GLM becomes very challenging with the increasing number of variables. As it is not possible to fit and compare all possible models for a large number of variables, three classical approaches have been developed: forward selection, backward selection, and mixed (bi-directional) selection. Forward selection is a greedy approach and might include variables early that later become redundant. Mixed selection can fix this challenge. As an alternative to stepwise variable selection, one can use penalized likelihood estimation, e.g., Least Absolute Selection and Shrinkage Operator (LASSO) introduced by \cite{tibshirani1996}, to find the best subset of variables for a GLM. 

However, it is even more computationally challenging to use the above methods to search for the best GLM with interacting variables, as the number of possible interactions is too large, whereas re-fitting even one single GLM on a real-world big data set is time-consuming.  Therefore, researchers explored the usage of neural-network based models for predicting claim frequencies and learning from them about the interacting variables, which can be added to a GLM to improve its performance.

First, a neural-network based model is trained. Second, interaction-detection methods are applied. Two types of interaction detection methods are distinguished --- model agnostic and model specific ones. Model-agnostic methods do not use a specific structure of the ML model. This class of methods includes Friedman H-statistics \cite{friedman2008}, Greenwell statistics \cite{greenwell2018}, feature interaction in terms of prediction performance \cite{oh2019}, and SHAP interaction values \cite{lundberg2017}. The main drawback of the above-mentioned model-agnostic methods is their high computational cost for big actuarial data sets.  Model-specific interaction detection methods rely on the peculiarities of the ML model under consideration. For example, \cite{wuethrich2020} propose a procedure for determining missing interactions in the benchmark GLM via CANNs. For each interaction of interest, a CANN that uses the interaction of interest and the prediction of the benchmark GLM is trained. If the deviance loss function of the CANN model decreases significantly in comparison to the deviance loss of the benchmark GLM, then this interaction is considered as missing. \cite{richman2021} propose a LocalGLMnet, which retains the additive decomposition of the response variable as in the case of a GLM, but lets the regression coefficients become feature-dependent. Once a LocalGLMnet is trained, one can detect whether there is an interaction between two features by exploring smoothed plots of gradients of the regression coefficients, also called regression attention due to their dependence on the features. How to optimally determine interactions for categorical variables with many levels is still an open question for LocalGLMnets.

In our approach, we also use CANN, as in \cite{schelldorfer2019}. We use embedding layers for categorical features with many categories, as it is shown to improve predictive performance on actuarial data sets, see \cite{schelldorfer2019}, \cite{wuethrich2020}. To extract interactions, we use a model-specific method that is a modification of a method called Neural Interaction Detection (NID). This method is developed in \cite{tsang2017detecting}. It computes the strengths of all interactions among input neurons very of a neural network very fast, since it uses only trained weights of the neural network, and is extendable to embedding layers and the architecture of a CANN  model.




\textbf{Structure.} In Section 2, we explain the basics of GLMs. In Section 3, we describe in detail our proposed algorithm for detecting the next-best interaction for a benchmark GLM. Each subsection within this section is devoted to a specific part of our algorithm. Section 4 contains case studies, where we apply the proposed algorithm to two data sets --- an artificially created one and an open-source one -- and briefly comment on its usage for big confidential data sets. We offer our conclusions in Section 5. In Appendix \ref{app:genetic_algorithms} we provide information on advanced fine-tuning of neural networks with the help of genetic algorithms. Appendix \ref{app:NID_code} contains \texttt{R}-code for the neural interaction detection algorithm.

\section{Generalized linear models for modeling insurance claim frequencies}
\label{sec:GLM}

In this section, we briefly describe the basics of a GLM for modeling claim counts. We start with a definition of a GLM without interacting variables and then explain how an interaction is added to a GLM. This section is mainly based on  \cite{wuethrich2020}. Since their introduction in 1972, GLMs have enjoyed great popularity for modeling and forecasting claim frequencies within the insurance sector. 


Let a data set be denoted by $\{(N_i, \pmb{x}_i, v_i)\}_{i=1}^n$, where $n\in \mathbb{N}$ is the number of observations, $v_i \in [0,1]$ corresponds to the exposure time in years of the $i$-th observation (the time length in which events occur), $N_i\in \mathbb{N} \cup \{0\}$ refers to the number of claims observed for the $i$-th observation within exposure time $v_i$, and $\pmb{x}_i \in \mathcal{X} \subset \{1\} \times \mathbb{R}^{p}$ represents the vector of variables (features, covariates) for the $i$-th observation excluding the exposure time\footnote{For each vector of variables, its first component is always $1$ and serves the purpose of modeling an intercept component of a GLM}, $p\in \mathbb{N}$ is the number of features. In a GLM context, $N_i, i = 1,\dots, n,$ are assumed to be independent random variables that follow the Poisson distribution. The mean of the distribution of $N_i$ is assumed to depend on the so-called linear (systematic) component $\eta(\pmb{x}_i) := \langle \pmb{\beta} , \pmb{x}_i \rangle = \pmb{\beta}^\top \pmb{x}_i$  and the exposure time $v_i$ as follows:
\[N_i \sim \text{Poisson}(v_i \exp(\eta(\pmb{x}_i)),\]
where $\pmb{\beta} \in \mathbb{R}^{p + 1}$ is the vector of GLM parameters.  

Denote $\lambda^{GLM}(\pmb{\beta}, \pmb{x}_i) := \exp(\eta(\pmb{x}_i))$. The vector of parameters $\pmb{\beta}$ is estimated via the maximum likelihood estimation method. We denote the estimated parameters by $\hat{\pmb{\beta}}$ and the estimated (expected) annual number of claims for an observation $\pmb{x}_i$ by $\hat{\lambda}^{GLM}_i := \lambda^{GLM}(\hat{\pmb{\beta}}, \pmb{x}_i)$. The weighted average predicted frequency (WAPF) and the weighted average observed frequency (WAOF) are then defined as
\begin{equation}
\text{WAPF} = \frac{\sum_{i} \hat{\lambda}^{GLM}_i v_i}{\sum_i v_i}, \text{    }\text{    } \text{WAOF} = \frac{\sum_{i} N_i}{\sum_i v_i}.
\label{eq:WAPF&WAOF}
\end{equation}

An important requirement for modeling claim counts is the equality of the WAPF and the WAOF. This property is called a balance property. A GLM satisfies this property on the data set used for model fitting. For the proof of this fact, an interested reader is referred to Equation (2.10) in \cite{wuethrich2019}.

In GLMs, different choices of variables lead to different predictive performance. The optimal set of variables can be found by fitting GLMs with different subsets of variables and comparing goodness-of-fit measures like the Akaike information criterion (AIC), the Bayesian information criterion (BIC), etc.


These model evaluation criteria can be used for the automated selection of the best subset of available variables for a GLM. Popular automated feature selection methods are stepwise backward, forward, and mixed variable selection methods. As mentioned in the introduction section, these methods have huge computational costs for data sets with a large number of variables. 


%
%

Next, we explain how to include a pairwise interaction to a GLM. Let $x_{\cdot,1}$ and $x_{\cdot,2}$ be two features whose pairwise interaction should be added to a GLM. It means that a parametric term $I(x_{\cdot,1},x_{\cdot,2})$ related to those features should be added to the linear (systematic) component $\eta(\pmb{x})$ of a GLM. The following cases are distinguished:
\begin{itemize}
    \item \textbf{Interaction between two numerical features} \\
    Let $x_{\cdot,1}$ and $x_{\cdot,2}$ be two numerical features. For an observation $i$, the term modeling the interaction\footnote{The is a simple parametric form of an interaction. In general, transformations of involved variables may be needed, e.g., raising to a power or taking a logarithm.} between them is defined by \[I(x_{i,1},x_{i,2}) = \beta_{1,2} \cdot x_{i,1} \cdot x_{i,2},\]
    where $\beta_{1,2}$ is a parameter to be estimated.
    \item \textbf{Interaction between one numerical and one categorical feature} \\
    Let $x_{\cdot,1}$ be a numerical feature and $x_{\cdot,2}$ be a categorical feature with $J$ categories, where the last one serves as a reference category (also called a base level). For an observation $i$, the interaction between them is defined by \[I(x_{i,1},x_{i,2}) = \sum_{j=1}^{J-1} \beta_j \cdot x_{i,1} \cdot \mathbbm{1}_{\{x_{i,2}=j\}},\] where $\beta_{j}$ are parameters to be estimated, $\mathbbm{1}_{\{x_{i,2}=j\}}=1$ if the $i$-th observation of feature $x_{\cdot,2}$ is a category $j$, $0$ otherwise. 
    \item \textbf{Interaction between two categorical features} \\
    Let $x_{\cdot,1}$ and $x_{\cdot,2}$ be two categorical features with $R$ and $S$ categories respectively, where the last one each serves as a reference category. For an observation $i$, the interaction between features $x_{\cdot,1}$ and $x_{\cdot,2}$ is modeled by \[I(x_{i,1},x_{i,2}) =  \sum_{r=1}^{R-1} \sum_{s=1}^{S-1} \cdot  \beta_{r,s} \cdot \mathbbm{1}_{\{x_{i,1}=r\}} \mathbbm{1}_{\{x_{i,2}=s\}},\] where $\beta_{r,s}$ are parameters to be estimated, $\mathbbm{1}_{\{x_{i,1}=r\}}=1$ if the $i$-th observation of feature $x_{\cdot,1}$ is a category $r$ and $0$ otherwise and likewise $\mathbbm{1}_{\{x_{i,2}=s\}}=1$ if the $i$-th observation of feature $x_{\cdot,2}$ is a category $s$ and $0$ otherwise.
\end{itemize}

The search of important interactions is more challenging than the search of the best subset of variables for a GLM, since the number of all possible combinations of interacting variables is usually larger than the number of variables\footnote{For $p$ variables, there are $p(p - 1) / 2$ possible pairwise interactions}. Therefore, actuaries often use their expert knowledge to decrease the number of pairwise interactions to analyze in detail. The interactions to be analyzed are mainly explored in a visual manner, e.g., by evaluating plots that indicate the (weighted) average of the response variable for each unique combination of values of variables (or their binned versions).

In the next section, we describe in detail our suggested approach to detecting important pairwise interactions. It is faster than the majority of methods proposed in the literature and, thus, may save actuaries time to focus on other challenging tasks.

\section{Algorithmic detection of the strongest interaction missing in a GLM}
\label{sec:AlgorithmicDetectionOfTheStrongestInteractionMissingInAGLM}



From now on we refer to the GLM that is to be improved as the benchmark GLM. To detect the next-best interaction for the benchmark GLM, we suggest an algorithm that consists of three steps:
\begin{enumerate}
    \item Outperform the benchmark GLM using a CANN model.
    \item Rank the interactions learned by the CANN according to their strength.
    \item Determine the most significant top-ranked interaction via mini-GLMs.
\end{enumerate}
We refer to the model developed in Step 1 as the competitor (ML) model. Note that CANNs and other machine-learning models cannot yet replace benchmark GLMs used by insurance companies in production for various reasons: lack of interpretability, etc.  

\subsection{Outperforming the benchmark GLM via CANN}
\label{subsec:OutperformingTheBenchmarkGLMViaRegularizedNeuralNetwork}

As previously mentioned, GLMs have been a traditional technique for modeling and forecasting claim frequencies within the insurance sector. However, according to \cite{wuethrich2019} their performance is limited in comparison to models based on NNs,which by their design learn non-linear interactions between variables. Thus, in Step 1 of the suggested approach, we train a CANN that can be seen as a boosting step for the benchmark GLM and is, essentially, a NN that uses the predictions of the benchmark GLM while learning additional interactions between variables to improve the predictive power of the overall model. Before explaining CANNs in more detail, we provide the basics of NNs. 

Since NNs do not satisfy the balance property\footnote{NNs violate the balance property due to early stopping that is used to prevent overfitting.}, we use them for transforming data (representation learning). After the representation of the original data set is learned, we use this transformed data to predict claim counts via a GLM. In this section, we briefly describe the above-mentioned modeling aspects. For more information, we refer the interested reader to \cite{ferrario2018insights} and \cite{wuethrich2020}.


Consider a fully-connected feed-forward NN with $d \in \mathbb{N}$ hidden layers and one neuron in the output layer. Denote by $q_l\in \mathbb{N}$ the number of neurons in the $l$-th hidden layer,  $l = 1, \dots, d$. $q_0\in \mathbb{N}$ denotes the number of neurons in the input layer. In our application of NNs, $q_0$ will be equal to the number of neurons needed to encode $p$ variables in the original data set and may be larger than $p$ since each categorical variable usually needs more than one neuron in the input layer. Denote by $\tilde{\pmb{x}} \in \mathbb{R}^{q_0}$ the vector of pre-processed features that serve as input to the NN. Denote by $W^{(l)} \in \mathbb{R}^{q_l \times q_{l-1}}$ the weight matrices and by $\pmb{b}^{(l)} \in \mathbb{R}^{q_l}$ the bias vectors, $l = 1, \dots, d$. Denote by $\pmb{w}^{y} \in \mathbb{R}^{q_d}$ and by $b^{y} \in \mathbb{R}$ the coefficients vector and bias for the output neuron. Denote by $\phi_l(\cdot)$ the activation function of neurons in the $l$-th layer $l = 1, \dots, d+1$ and by $\overrightarrow{\phi_l} (\pmb{\xi})  = (\phi_l(\xi_1), \dots, \phi_l(\xi_{q_l}))^\top$ for any $\pmb{\xi} \in \mathbb{R}^{q_l}$. Then the hidden layers $\pmb{z}^{(l)}$ and the output layer consisting of one neuron $y$ (the NN's prediction) can be expressed as follows: 
\begin{equation*}
    y = \phi_{d+1}\left( (\pmb{w}^y)^\top \pmb{z}^{(d)} + b^y\right), \quad \pmb{z}^{(l)} = \overrightarrow{\phi_l}\left( W^{(l)}\pmb{z}^{(l-1)} + \pmb{b}^{(l)} \right), \quad l = 1,\dots, d,
\end{equation*}
with $\pmb{z}^{(0)} := \tilde{\pmb{x}}$. Let $\phi_{d+1}(z) = z$ and denote the regression function of a NN by
\begin{equation}\label{eq:NN_regression}
    \lambda^{\text{NN}}(\xtil):= \rBrackets{w^y}^\top  \rBrackets{\pmb{z}^{(d)} \circ \pmb{z}^{(d - 1)} \circ \dots \circ \pmb{z}^{(1)}}(\xtil) + b^y.
\end{equation}

A combined actuarial neural network (CANN) for claim counts satisfies two model assumptions:
\begin{enumerate}
    \item $N_i \sim \text{Poisson}(v_i \cdot \lambda^{\text{CANN}}(\tilde{\pmb{x}}_i))$ with the regression function $\lambda$ given by
    \begin{equation}\label{eq:CANN_regression}
        \tilde{\pmb{x}}_i \mapsto \ln \rBrackets{\lambda^{\text{CANN}}(\xtil_i) } = \ln(\hat{\lambda}^{\text{GLM}}_i) + \lambda^{\text{NN}}\rBrackets{\xtil_i}
    \end{equation}
    \item The regression function in \eqref{eq:CANN_regression} is initialized with weights $w^y = (0, 0, \dots, 0)^\top \in \mathbb{R}^{q_{d}}$ and $b^y = 0$.
\end{enumerate}

The first structural assumption means that the NN part of CANN boosts the benchmark GLM. The second structural assumption implies that at the beginning of the training phase, the Poisson deviance of a CANN model equals the Poisson deviance of the benchmark GLM. If the Poisson deviance loss is used as an objective function for training CANN, then during training the gradient-descent algorithm explores the NN architecture for additional model structure that is not present in the benchmark GLM and that further decreases the  CANN's Poisson deviance.

It is not necessary to know the structure of the benchmark GLM, only its predictions are used by a CANN model. So the weights of the benchmark GLM are non-trainable and the implementation of a Poisson CANN can be simplified by merging the annualized predictions of the benchmark GLM with the given volumes. In particular, as an alternative to the first structural assumption of a Poisson CANN, we can consider $N_i \sim \text{Poisson}(v^{\text{GLM}}_i \cdot \lambda^{\text{NN}}(\tilde{\pmb{x}}_i))$ with modified exposure $v_i^{\text{GLM}}:= v_i \cdot \hat{\lambda}^{\text{GLM}}_i$.

\begin{figure}[H]
    \centering
  \includegraphics[width=\textwidth]{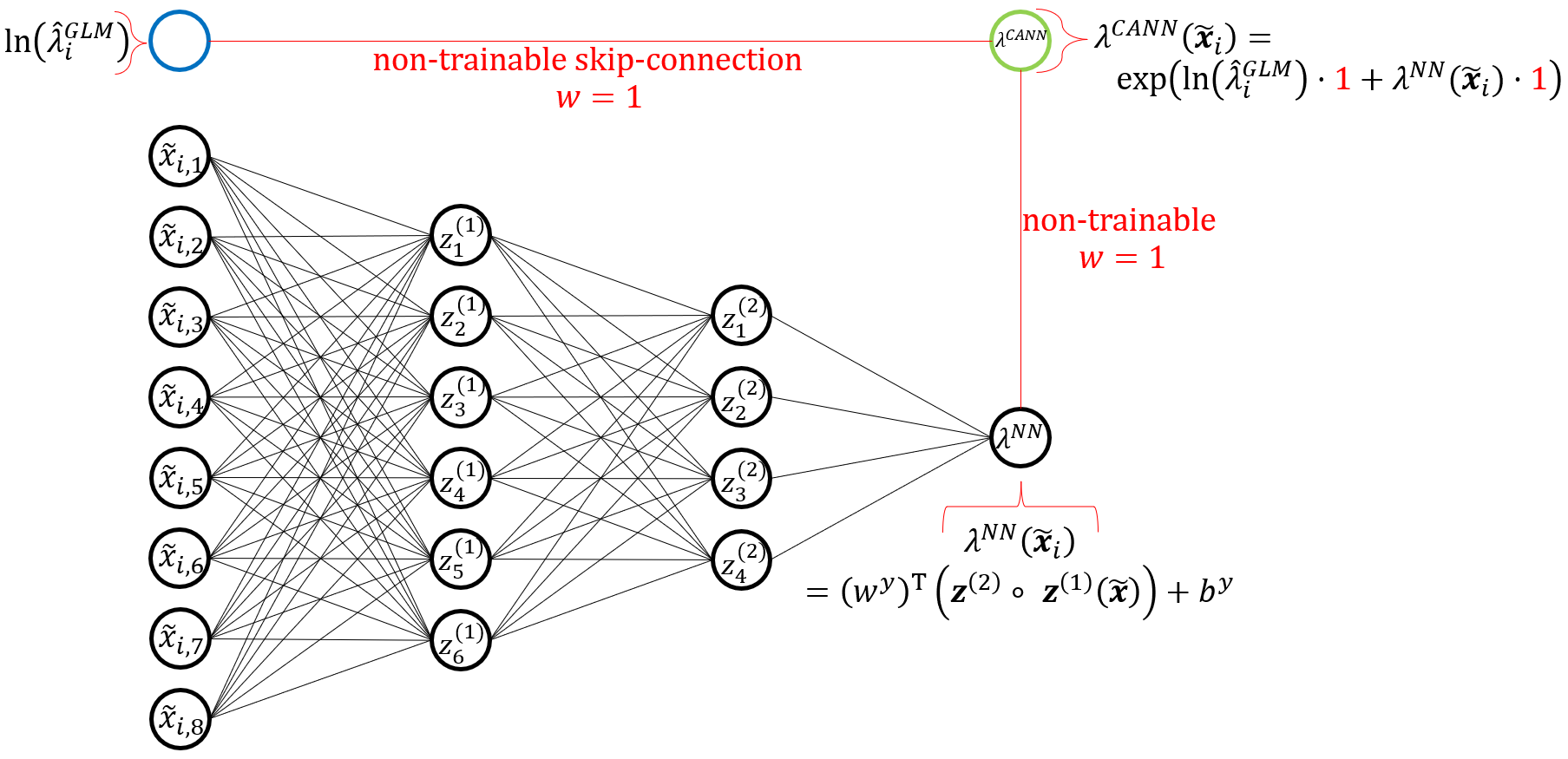}
  \caption{Architecture of a CANN model} 
  \label{fig:CANN_example}
\end{figure}

The architecture of a CANN is illustrated in Figure \ref{fig:CANN_example}. The neuron marked blue takes as input the modified exposure $v_i^{\text{GLM}}$ and passes it directly to the output neuron marked green, whereas the corresponding red-marked connection has a non-trainable weight $1$ and a bias coefficient $0$. The neurons marked black and connections among them constitute the NN component of a CANN. In the NN component of the CANN model shown in Figure \ref{fig:CANN_example}, $q_0 = 8, q_1 = 6, q_2 = 4, q_1 = 1$. The output $\lambda^{\text{NN}}$ of the NN component is passed via the red-marked connection (with a non-trainable weight $1$) to the green-marked output neuron. The output neuron sums two incoming values and applies the exponential activation function on the result. {The CANN model is trained using the Poisson deviance loss function.} 

As mentioned above, the original vector of variables $\pmb{x} \in \mathbb{R}^{p}$ does not enter the input layer of a NN, but its pre-processed version $\tilde{\pmb{x}}\in \mathbb{R}^{q_0}$ does. In particular, all features that appear in the input layer of a NN, must not contain missing values, should be numerical, and ideally have the same range. Therefore, we use min-max-scaling to all numerical features used for training a NN. As for categorical features, we recommend using one-hot encoding for features with a low number of unique categories, e.g., below $5$, and use the embedding layers technique for those with a larger number of unique categories. These techniques are recommended in \cite{wuethrich2020}.

\begin{figure}[H]
    \centering
  \includegraphics[width=\textwidth]{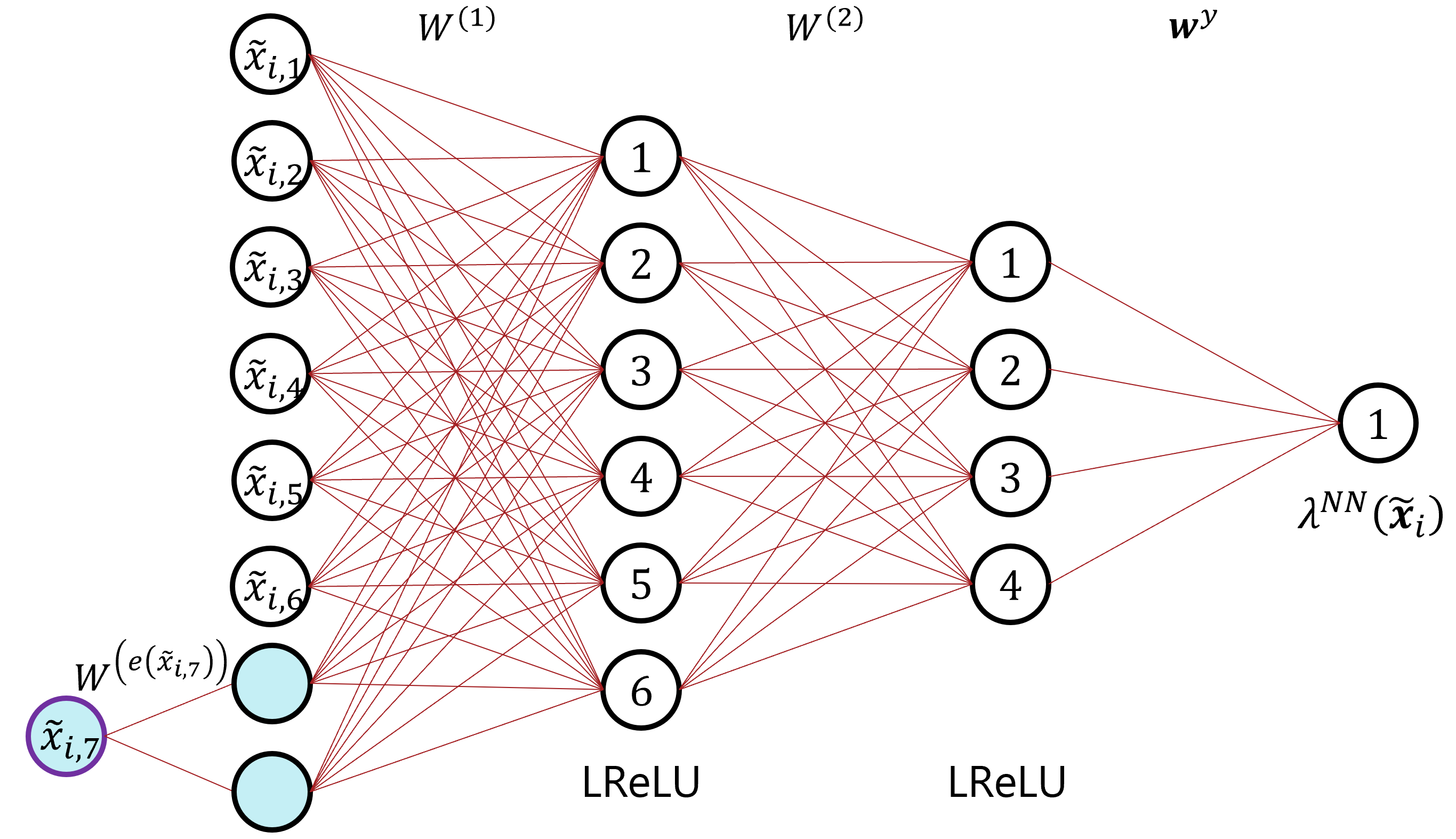}
  \caption{Example of the NN part of a CANN model that uses a $2$-dimensional embedding layer (in light blue) encoding a categorical feature $\tilde{x}_{\cdot, 7}$.} 
  \label{fig:EmbeddingLayer}
\end{figure}

An embedding $e$ of a categorical feature with $k$ distinct categories $\{a_1,...,a_k\}$ is a mapping
\begin{equation*}
    e:\{a_1,\dots,a_k\}\rightarrow \mathbb{R}^{g},   \text{   } a\mapsto e(a),
\end{equation*}
with $g\in \mathbb{N}$ denoting the dimension of the embedding. This dimensionality parameter is chosen by the user, whereby typically $g \ll k$.
The components \[e(1)_1,\dots,e(1)_g,\dots, e(k)_1,\dots,e(k)_g\] of such an embedding of $k$ categories constitute additional NN weights that are learned during training. So an embedding layer of dimension $g$ results in additional $g\cdot k$ embedding weights. The embedding representation of an embedded feature, i.e., the output of the embedding layer equals the embedding weights. A NN with the above-mentioned peculiarities is schematically illustrated in Figure \ref{fig:EmbeddingLayer}.

To reduce the risk of overfitting, we recommend using a dropout technique and an early stopping of the NN training process. According to the drop-out technique, a pre-specified percentage (a so-called dropout rate) of neurons randomly selected in each layer is \enquote{switched off} and not updated in a NN training step. According to the early-stopping method, the NN training is stopped as soon as a significant deterioration or no significant improvement in the model performance is observed within a predefined period of time. 

A Poisson CANN model does not satisfy the balance property \eqref{eq:WAPF&WAOF}. As we will see in the numerical studies and as it is also found in the numerical studies of \cite{schelldorfer2019}, the violation of the balance property is very small, since a CANN model uses the predictions of a GLM that fulfills this property. In the view of the interaction detection as the main focus of our paper, the violation of the balance property is negligible. Readers interested in enforcing the balance property on neural networks are referred to \cite{wuethrich2019} and \cite{wuethrich2020}.




Before training the above-described CANN, one has to specify certain hyper-parameters, e.g., the embedding dimension, the number of hidden layers, the number of neurons per layer, the dropout rates, the activation functions, loss functions and the optimizer, the batch size, the number of epochs, or the usage of early stopping, etc. Our experiments on both artificial and real data sets show that choosing 3 hidden layers, an embedding dimension of 2, a Poisson deviance as a loss function, and a dropout rate of $20\%$ has very high chances for a NN to outperform sophisticated benchmark GLMs. If one would like to further improve further the performance of the ML model, then one should explore ML models with different values of hyper-parameters. The search for the optimal values of hyper-parameters can be done either via a grid search or a genetic algorithm. The latter approach is more time-consuming, but it can yield a better model performance. More information on the genetic approach to optimizing hyper-parameters of NNs can be found in Appendix \ref{app:genetic_algorithms}.

To compare the performance of a ML model (for short competitor) and a benchmark GLM (for short benchmark) in the context of MTPL insurance claim counts, we recommend using so-called double lift plots on the test data set, which are of high practical importance for actuaries, see, e.g., Section 7.2.2 in \cite{goldburd2017}. A double lift-plot requires predictions of each of the two models and the true observed values of the response variable.
A double lift plot is created in the following way:
\begin{enumerate}
    \item Determine the deviance $\delta_i$ (also called the sort ratio), which is the relative difference between the competitor model and the benchmark GLM:
    \[\delta_i = \frac{\hat{\lambda}^{\text{competitor}}_i}{\hat{\lambda}^{\text{benchmark}}_i} - 1,\]
    where $\hat{\lambda}^{competitor}_i$ denotes the $i$-th prediction of the competitor model and $\hat{\lambda}_i^{\text{benchmark}}$ refers to the $i$-th prediction of the benchmark GLM.
    \item Sort the observations based on $\delta_i$, from smallest to largest.
    \item Bucket the observations into predetermined bins in an interval of interest, e.g., bins $(-\infty, -0.5]$, $(-0.5, -0.48]$, $(-0.48, -0.46], \dots, (0.48, 0.5]$,  $(0.5, +\infty)$.
    \item For each bin, calculate the exposure, WAOF, WAPF of the competitor model, and WAPF of the benchmark model.
    \item For each bin, plot the quantities calculated in Step 4. The left y-axis refers to the WAOF or WAPF that are marked by dots in the double lift plot. The right y-axis refers to the exposure that is depicted by bars below the dots. 
\end{enumerate}

\begin{figure}[H]
    \centering
  \includegraphics[width=\textwidth]{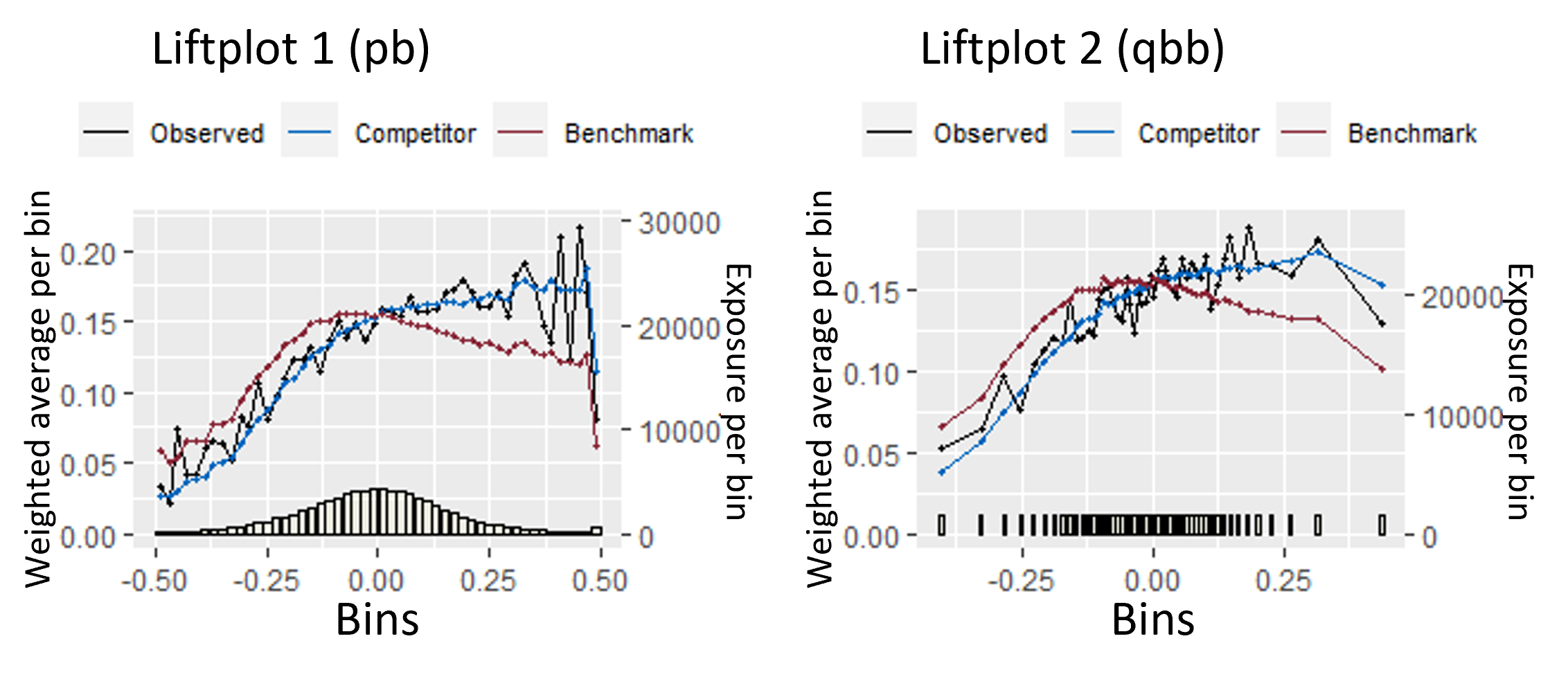}
  \caption{Lift plots.} 
  \label{fig:LiftPlots}
\end{figure}

An example of a lift plot with predetermined binning can be seen in the left sub-figure of Figure \ref{fig:LiftPlots}. As an alternative to bucketing the observations based on the predetermined binning, one can use quantile-based binning. A double-lift plot of this type can be seen in the right sub-figure of Figure \ref{fig:LiftPlots}. In this chart, each bin has the same number of observations and is determined based on the quantiles of the distribution of $\delta_i$.

\begin{itemize}
    \item Lift plot 1 (pb):  3 curves (observed, competitor, benchmark), predetermined bins
    \item Lift plot 2 (qbb):  3 curves (observed, competitor, benchmark), quantile-based bins
\end{itemize}

Obviously, evaluating these lift plots is based on visual perception. In order to allow for a purely quantitative model evaluation, we construct KPIs reflecting the information captured in these lift plots and, thus, not requiring visual evaluation of lift plots. 

Let $B=\{1,...,|B|\}$ be the set of bins in a lift plot and $b \in B$ an index of a certain bin. Define the weighted exposure $w_b$ per bin as follows:
\begin{equation*}
    u_{b}=\frac{\sum_{x_{i} \in \mathcal{X}_{b}} v_{i}}{\sum_{x_{i} \in \mathcal{X}} v_{i}},
\end{equation*}
where $\mathcal{X}_b$ is the set of features vectors that correspond to observations in bin $b \in B$. The numerator equals the total exposure of observations in a bin $b$ and the denominator is equal to the total exposure in the whole data set used for calculating this KPI. Using this weighted exposure per bin, the mean absolute error based on the lift plot bins is given by
\[\texttt{mae\_lift\_...}= \sum_{b \in B} u_{b}|\text{WAPF}_{b}-\text{WAOF}_{b}|.\]
All in all, we thereby construct a selection of numerical KPIs, namely
\begin{itemize}
    \item KPI lift plot 1:  \texttt{mae\_lift\_pb(\_benchmark)}  (mean absolute error based on the lift-plot with predetermined bins and benchmark GLM)
    \item KPI lift plot 2:  \texttt{mae\_lift\_qbb(\_benchmark)}  (mean absolute error based on the lift-plot with quantile-based bins and benchmark GLM)
\end{itemize}

The \texttt{mae\_lift\_\dots} KPIs represent the mean absolute error of the model to be compared with the benchmark GLM, whereas \texttt{mae\_lift\_\dots\_benchmark} refers to the mean absolute error of the benchmark GLM. The smaller the value of \texttt{mae\_lift\_\dots}, the better the model. For example, if \texttt{mae\_lift\_pb} is smaller than  \texttt{mae\_lift\_pb\_benchmark}, the competitor model outperforms the benchmark model based on the lift-plot with predetermined bins. The same reasoning holds for the KPIs using quantile-based binning. 

In summary, actuaries often rely on lift plots when evaluating model performance. However, the visual interpretation of such lift plots may be rather subjective. Hence, the transformation of the lift plot into a numeric KPI and using it along Poisson deviance for model selection may enhance reliability and the objectivity of the performance evaluation. For example, it may happen\footnote{We have observed such a situation when applying our methodology to a big proprietary data set} that Poisson deviance is the same for two models, but one model is convincingly better than the other one according to lift-plot based KPIs.


\subsection{Opening the black box: ranking learned interactions}
\label{subsec:OpeningTheBlackBox:RankingLearnedInteractions}$\,$

Having found a well-performing CANN, the next step is to find the most significant pairs of interacting variables learned by the model. Here, \enquote{significant} means that those pairs of interacting variables that are captured by the CANN model are likely to strongly improve the predictive power of the benchmark GLM if included in it. In the CANN model, its NN component learns non-linear interactions among the input features. To quantify the significance of each of the learned interactions, we apply a fast model-specific interaction-detection method. The method can be seen as an adjustment of a technique called Neural Interaction Detection (NID), proposed by \cite{tsang2017detecting} for fully-connected feed-forward NNs.

\begin{figure}[H]
    \centering
  \includegraphics[width=0.8\textwidth]{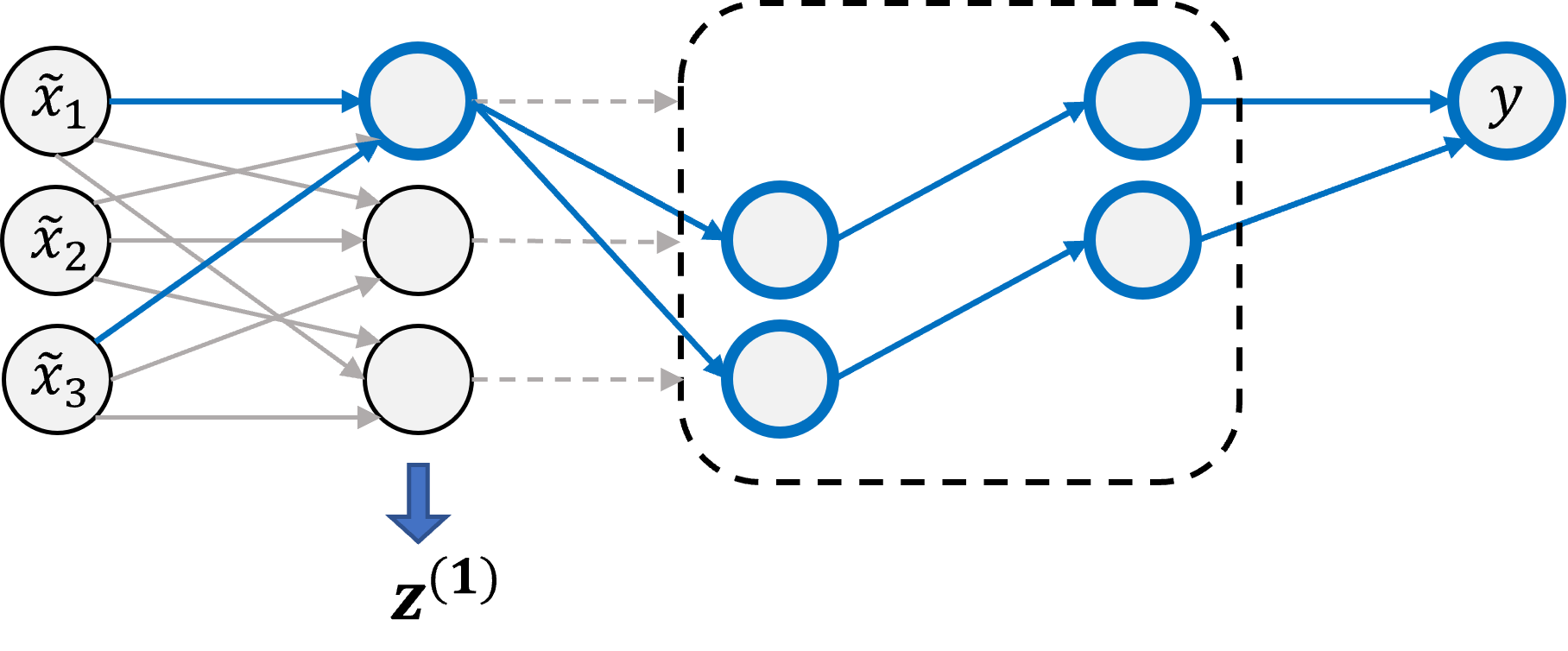}
  \caption{Generation of interactions in the first hidden layer and propagation of these interactions through the network. Figure adapted from \cite{tsang2017detecting}.} 
  \label{fig:NID_1}
\end{figure}

The original NID algorithm is based on the assumption that feature interactions are created in the first hidden layer of a neural network. Note that learning interactions in the first hidden layer is possible due to the usage of non-linear activation functions. Moreover, \cite{tsang2017detecting} provide empirical evidence that considering the first hidden layer is indeed sufficient for determining interactions. These interactions are then propagated through the whole network and influence the final prediction. This concept is exemplary shown in Figure \ref{fig:NID_1}. As can be seen, the first neuron in the first hidden layer $\pmb{z}^{(1)}$ (highlighted in blue) takes inputs $\tilde{x}_1$ and $\tilde{x}_3$ and thereby creates an interaction between them if the activation function of that neuron is non-linear. The strength of this interaction is evaluated based on both incoming weights as well as the outgoing paths from the neuron to the output neuron $y$ as colored in blue in Figure \ref{fig:NID_1}. The higher the incoming weights and the higher the impact of the considered neuron on the final output, the stronger the interaction. The strength of an interaction is quantified by an interaction strength score.

Let $I$ be a pair of input neurons. The interaction between these input neurons happens at each neuron of the first hidden layer. Denote by $s_j(I)$ the strength of an interaction between input neurons in $I$ measured at the $j$-th neuron in the first hidden layer, $j \in q_1$. It is quantified as follows:
\begin{equation}
s_j(I)=\zeta_j^{(1)}\cdot\mu(|W_{j,I}^{(1)}|), \text{ } s_j(I) \in \mathbb{R},
\label{eq:NID_interaction}
\end{equation}
where $\zeta_j^{(1)}$ represents the influence of neuron $j$ on the model prediction, $|W_{j,I}|$ denotes the absolute value of the incoming weights from features in $I$ to neuron $j$ in the first hidden layer, and $\mu(\cdot)$ represents a so-called generalized surrogate function used to capture the strength of the interaction based on the relevant incoming weights. In our notation, $|\cdot|$ applied to a matrix means that the absolute value is taken element-wise, i.e., for all matrix elements.

As per \cite{tsang2017detecting}, the generalized surrogate function $\mu(\cdot)$ should be such that interaction strength is
\begin{enumerate}
    \item quantified as zero when the interaction does not exist;
    \item non-decreasing in the magnitude of feature weights;
    \item less sensitive to changes in large feature weights.
\end{enumerate} 
The third property mitigates the impact of situations, when the weight of the connection from one input neuron has much higher magnitude than the weight of connection from another input neuron. If the large weight grows in magnitude, then interaction strength should not change much, but if instead the smaller (in magnitude) weight grows at the same rate, then interaction strength should increase. Thus, maximum, root mean square and arithmetic mean are not suitable candidates for $\mu(\cdot)$. \cite{tsang2017detecting} empirically investigate a selection of possible surrogate functions and conclude that the minimum is the best-performing function that recovers the highest number of true interactions in their experiments. The second-best choice for $\mu(\cdot)$ was the harmonic mean function. Therefore, we choose $\mu(\cdot)$ as minimum in all our experiments.

The influence $\pmb{\zeta}^{(1)}$ on the network prediction is calculated as the following matrix product of the absolute weight matrices:
\begin{equation}
\pmb{\zeta}^{(1)}=|\pmb{w}^{y}|^{\top}\cdot|W^{(d)}|\cdot|W^{(d-1)}|\cdot ... \cdot|W^{(2)}|, \text{ } \pmb{\zeta}^{(1)} \in \mathbb{R}^{q_1}.
\label{eq:NID_influence}
\end{equation}
In this case, $q_1$ denotes the number of neurons of the first hidden layer, $W^{(m)}$ each represents the weight matrix connecting the units between hidden layers $m-1$ and $m$, whereas $\pmb{w}^{y}$ denotes the vector of weights connecting the last hidden layer and the output neuron.
Note that $\pmb{\zeta}^{(1)}$ results in a vector where the $j$-th index corresponds to the influence of a neuron $j$ of the 1-st hidden layer on the output neuron of a NN. According to Lemma 3 in \cite{tsang2017detecting}, if all activation functions in a NN are $1$-Lipschitz continuous, then Definition \eqref{eq:NID_influence} is an upper bound for the gradient magnitudes of neurons in the first hidden layer, i.e., if $\left|\frac{\partial \phi(x) }{ \partial x} \right| \leq 1$, then  $\left|\frac{\partial y }{ \partial z_j^{(1)}} \right| \leq \zeta_j^{(1)}$ for all $j = 1, \dots, q_1$. Common activation functions such as rectified linear unit, hyperbolic tangent and sigmoid are $1$-Lipschitz continuous.



\begin{figure}[H]
\centering
  \includegraphics[width=0.8\textwidth]{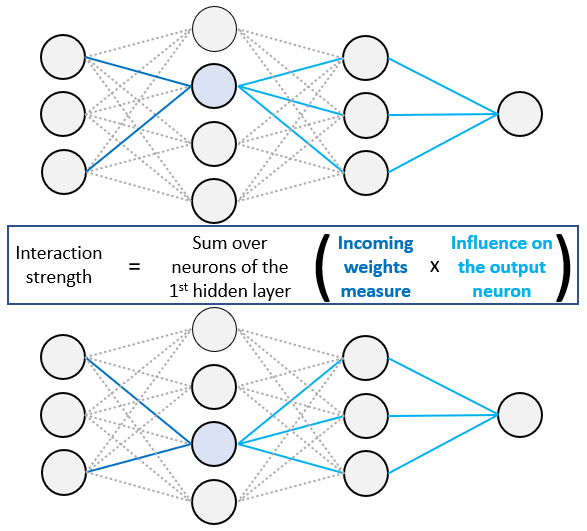}
  \caption{Illustration of the interaction-strength calculation: evaluate all neurons of the first hidden layer by measuring the in-going and outgoing paths and then aggregate the results.} 
  \label{fig:NN_interactions}
\end{figure}

After having extracted the incoming weights of the NN as well as the importance (w.r.t. the influence on the NN's output) of each neuron in the first hidden layer, the strength of a (local) interaction between a subset of input neurons can be computed for each neuron of the first hidden layer. Subsequently, the final interaction strength score for this subset of features is equal to the sum of local interaction strength scores across all $q_1$ neurons in the first hidden layer:
\[s(I) = \sum_{j=1}^{q_1} s_j(I),\]
which is exemplary illustrated in Figure \ref{fig:NN_interactions}.

Recall that a CANN model is a combination of a NN and a benchmark-GLM prediction with a skip-connection to the output neuron of the model, where the NN component is a feed-forward fully-connected neural network. Therefore, we can apply NID to the NN component of CANN model to quantify the strength of all pairwise\footnote{The strengths of interactions of the order higher than $2$ can be also computed via the NID approach. In that case, one can use a greedy approach to speed up calculations. For more information, see Section 4.2 in \cite{tsang2017detecting}.} interactions among features. {We provide the R-code for the described NID approach in Listing \ref{lst:NID} in Appendix \ref{app:NID_code}.}




The original NID method evaluates the strengths of interactions among input neurons. However, the encoding of categorical features may require several neurons in the input layer of a NN, e.g., in case of one-hot encoding one needs the same number of input neurons as the number of categories. If am actuary wants to detect interactions on a per-category level, then it is not neccessary to aggregate NID scores related to neurons encoding categorical variable. However, to obtain the interaction-strength scores related to a categorical feature taken as a whole, one has to aggregate the scores on a per-neuron basis using some aggregation function like mean, minimum, maximum. If an actuary is interested in finding interactions where the majority of categories of the categorical feature of interest are strongly interacting with another variable, then $min$ is recommended. If the aim is to find categorical variables whose categories have on average high interaction-strength scores with the other variable, then $mean$ is a good choice for aggregation. Finally, if the an actuary is interested in finding categorical variables where one category is especially strongly interacting with the other variable, then we recommend to use $max$ as the aggregation function. For example, to get the strength of an interaction between a categorical one-hot encoded feature and a numerical feature, one can take the maximum of all $s(I)$ where $I$ contains a input neuron encoding a category of the categorical feature of interest and an input neuron encoding the numerical feature of interest.

\subsection{Identification of the next-best interaction for a GLM}
\label{subsec:IdentificationOfTheNextBestInteractionForAGLM}

After extracting the most significant interactions, the final step is to determine the next-best interaction for the benchmark GLM. This step is necessary for several reasons. First, the inclusion of any interaction in a GLM requires a parametric specification of the interaction. This is also important for preserving the interpretability of the benchmark GLM. Second, it may happen that several top-ranked interactions have very similar interaction-strength scores according to NID, which is why choosing the next-best interaction may become ambiguous. In this case, an actuary may want to estimate the improvement of the benchmark GLM for each of the top-ranked interactions and afterwards decide which to include and retrain the benchmark model with the found interaction.

To decide which interaction to add to the benchmark GLM, we suggest to predict the observed claim counts via \enquote{mini} GLMs that use the predictions of the benchmark GLM and the top-ranked interactions. This approach can also be interpreted as freezing the coefficients of the benchmark GLM and adding one interaction on top to better predict the claim counts. The approach works as follows:



\begin{enumerate}
    \item For each $(x_{\cdot, j}, x_{\cdot, k})$ from the list of top-ranked interactions $\mathcal{I}_{top}$ and for each {relevant} parametric form of $I(\cdot, \cdot)$  
    \begin{enumerate}
        \item Fit a mini-GLM:
        \begin{equation*}
            N_{\cdot} \sim \text{Poisson}(v_{\cdot}\hat{\lambda}^\text{benchmark}_{\cdot} \cdot e^{I(x_{\cdot,j}, x_{\cdot,k})}),
        \end{equation*}
        \item Calculate KPIs of interest, e.g., AIC, residual deviance, etc.
    \end{enumerate}
    \item Recommend as the next-best interaction the one that corresponds to the mini-GLM with the best KPI.
\end{enumerate}

\textbf{Remarks}
\begin{enumerate}
    \item The word \enquote{relevant} refers to the fact that the exact form of the interaction $I(x_{\cdot,j}, x_{\cdot,k})$  depends on the types of features $x_{\cdot,j}$ and $x_{\cdot,k}$, as discussed in Section \ref{sec:GLM}.
    \item If at least one of the interacting variables is continuous, one has multiple options for choosing the parametric form of the interaction:
    \begin{enumerate}
        \item Consider several continuous transformations of the continuous feature(s) of interest. For example, an actuary may consider only parametric interactions of the form of $I(x_{\cdot,j}, x_{\cdot,k}) = x_{\cdot,j}^{a} \cdot x_{\cdot,k}^b$ for $a \in \{ 1,2,3 \}$ and $b \in  \{ 1,2,3 \}$. The form that leads to a mini-GLM with the best KPI is chosen.
        \item Bin the continuous feature(s) of interest and include the interaction between the binned versions of those features. A simple binning procedure can be based on the quantiles of their distribution. A more advanced binning procedure can be based on fitting a generalized additive model (GAM) that uses only a smooth version of the interaction of interest and the predictions of the benchmark-GLM as offset. Afterwards one trains a regression tree that predicts the GAM-captured interaction effect using the interacting features and concludes the optimal binning from the splits of the regression tree. For more information on this method, see Section 4.2 in \cite{henckaerts2018}.
    \end{enumerate} 
    \item It may be computationally challenging\footnote{Some versions of the actuarial software Emblem have a technical limit on the number of categories in categorical variables used in a GLM. In those cases, clustering must be performed with the number of clusters below that technical limit.} to fit a mini-GLM for categorical features with a large number of categories, e.g., postcode. For these cases, we recommend clustering categories of such variables based on the embedded representations of those variables. In the numerical studies, we use the k-means clustering algorithm and the Calinski-Harabasz clustering-validation measure to determine the optimal number of clusters. For more information on these methods, we refer the interested readers to \cite{murphy2013} and \cite{calinski1974}.
\end{enumerate}

\section{Case studies}

In this section, we summarize the results of several case studies, which we conduct on a computer with 11th Gen Intel(R) Core(TM) i7-1185G7 @ $3.00$GHz processor, $32$ GB RAM, Intel(R) Iris(R) Xe Graphics, and Windows 10 Enterprise operating system. In the first case study, we work with an artificially generated data set, where we know the true interactions among variables in the data set. The aim of this case study is to show that our methodology detects and recommends the true interaction. In the second case study, we work with an open-source data set, where the true interactions in the data are not known. In the third case study, we briefly discuss the benefits of our methodology for big data sets, since big insurers have millions of observations and keep track of tens of variables.

\subsection{Artificial data set}\label{subsection:aritificial_data}

In this subsection, we apply the previously described methodology to an artificially created data set. We start with generating $2$ million vectors of features $\pmb{x} = (x_1, \dots, x_{10}) \in \mathbb{R}^{10}$. The first $8$ features are numerical and come from a multivariate normal distribution with zero mean and unit variance, as in \cite{richman2021}. We assume that all numerical features are independent except for $x_2$ and $x_8$, which have a correlation of $0.5$ Hence, we randomly generate $(x_1, \dots, x_8)^\top \sim N(0,\Sigma)$ with $\Sigma$ being an identity matrix with an additional entry of $0.5$ in the cells $(2,8)$ and $(8,2)$. The last $2$ features are categorical and come  from a binomial distribution. The feature $x_9 \sim \text{Binomial}(2, 0.3)$ has three categories $\{0, 1, 2\}$ and is independent of other features. The feature $x_{10} \sim \text{Binomial}(5, 0.2)$ has six categories $\{0, 1, 2, 3, 4, 5\}$ and is independent of other features. For simplicity, we assume that $v_i = 1,\, i = 1,\dots, 2 \cdot 10^{6}$.

Based on the above-generated features and exposure, we generate the claim counts as follows. First, we calculate
\begin{align*}
        \pmb{x}\in \mathbb{R}^{10} \mapsto \mu(\pmb{x})  =  & \exp \bigl(-3 + 0.5\cdot x_1 - 0.25\cdot x_2^2 + 0.5\cdot |x_3|\cdot \sin (2\cdot x_3) + {\color{blue} 0.5\cdot x_4 x_5}\\
        &+ {\color{blue} 0.125\cdot x_5^2 x_6} - 0.1 \cdot 1_{\{x_9 = 1\}} - 0.2 \cdot 1_{\{x_9 = 2\}} +
        0.1 \cdot 1_{\{x_{10}=1\}} \\
        &+ 0.2 \cdot 1_{\{x_{10}=2\}} + 0.3 \cdot 1_{\{x_{10}=3\}} +
        0.4 \cdot 1_{\{x_{10}=4\}} + 0.5 \cdot 1_{\{x_{10}=5\}}\bigr).
\end{align*}

For a small number of feature vectors it holds $\mu(\pmb{x}) > 1$. In those cases we set $\mu(\pmb{x}) = 1$ to avoid unrealistically large number of claim counts for those vectors. In the final step of the data generation process, we obtain claim counts by generating them as follows:

\begin{equation*}
    N_i \sim \text{Poisson}(\exp(\mu(\pmb{x}_i))), \quad i = 1,\dots, 2 \cdot 10^{6},
\end{equation*}

The structure of the resulting data set is summarized in Listing \ref{lst:Artificial_DataSetStructure}. 

\begin{lstlisting}[caption={Structure of the artificial data set}, label={lst:Artificial_DataSetStructure}]
'data.frame':	2000000 obs. of  14 variables:
 $ claim_total_nb : int  0 0 0 1 0 1 0 0 0 0 ...
 $ annual_exposure: num  1 1 1 1 1 1 1 1 1 1 ...
 $ x_1  : num  -0.185 1.131 1.305 1.049 0.355 ...
 $ x_2  : num  0.465 0.444 -1.664 -1.004 0.99 ...
 $ x_3  : num  0.2259 -0.4888 0.0332 0.2362 1.0171 ...
 $ x_4  : num  0.696 -0.537 -3.043 1.849 -1.085 ...
 $ x_5  : num  1.932 0.276 -0.41 1.086 1.257 ...
 $ x_6  : num  0.716 0.11 -0.214 1.137 0.63 ...
 $ x_7  : num  -0.533 -1.466 -0.276 -1.457 -1.446 ...
 $ x_8  : num  0.5058 -0.0455 -1.0353 0.8823 -1.2143 ...
 $ x_9  : Factor w/ 3 levels "0","1","2": 1 3 1 2 2 1 1 2 1 1 ...
 $ x_10 : Factor w/ 6 levels "0","1","2","3",..: 2 2 1 1 3 3 2 ...
\end{lstlisting}

We split the data set as follows: $80\%$ for training, $10\%$ for validation, $10\%$ for testing. This is a rule for splitting data according to \cite{burkov2021}. The training set is used for fitting the model, the validation set is utilized for fine-tuning the hyper-parameters of the ML model, and the test set is used for evaluating the final out-of-sample performance of the chosen best-performing ML model. This results in the claim distributions shown in Table \ref{tab:Artificial_ClaimDistribution}.

\begin{table}[H]
 \caption{Claim distribution}
 \label{tab:Artificial_ClaimDistribution}
 \begin{tabular*}{\textwidth}{llllllllll}
    \hline\noalign{\smallskip}
      && \multicolumn{7}{c}{number of observed claims} & claim \\
      && 0 & 1 & 2 & 3 & 4 & 5 & 6 & total\\
    \noalign{\smallskip}\hline\noalign{\smallskip}
    \multirow{2}{*}{Full} & num. &  1887159 & 105560 & 6516 & 645 & 103 & 15 & 2 & 112841\\
    & \% & 94.3580 & 5.2780 & 0.3258 & 0.0323 & 0.0052 & 0.0075  & 0.0001 & 100.00\\ \noalign{\smallskip}\hline 
    \multirow{2}{*}{Train} & num. & 1509486 & 84518 & 5250 & 540 & 88 & 11 & 2 & 90409\\ 
    & \% & 94.3491 & 5.28272 & 0.3281 & 0.0337 & 0.0055 & 0.0007 & 0.0001 & 80.12\\ \noalign{\smallskip}\hline 
    \multirow{2}{*}{Val.} & num. & 189254 & 10441 & 630 & 56 & 7 & 1 & 0 & 11135\\
    & \% & 94.4433 & 5.2104 & 0.3144 & 0.0279 & 0.0035 & 0.0005 & 0.0000 & 9.87\\ \noalign{\smallskip}\hline  
    \multirow{2}{*}{Test} & num. & 188419 & 10601 & 636 & 49 & 8 & 3 & 0 & 11297\\
    & \% & 94.3435 & 5.3080 & 0.3185 & 0.0245 & 0.0040 & 0.0015 & 0.0000 & 10.01\\
    \noalign{\smallskip}\hline
 \end{tabular*}
\end{table}

To fit a benchmark GLM, we use both training and validation data. In this GLM, we include terms $x_1$, $x_2^2$, $x_3$, $x_3^2$, $x_9$, $x_{10}$, which appeared in the data generation process. However, we do not include in the benchmark GLM the interactions between features $x_4$ and $x_5$ and between features $x_5$ and $x_6$, which are the true interactions according to the process of the artificial data generation. If our interaction-detection methodology works correctly, one of these interactions will be recommended as the next-best one to be included to the benchmark GLM.

\subsubsection{Step 1: Training CANN}
\label{subsubsec:Step1:Artificial_OutperformingTheBenchmark}

We conduct the following data pre-processing steps prior to training the CANN model:
\begin{itemize}
    \item Use one-hot encoding for the categorical feature $x_9$.
    \item Use a $2$-dimensional embedding layer for the categorical feature $x_{10}$.
    \item Apply min-max scaling to all numerical features $x_1, \dots, x_8$:
    \begin{equation*}
        \tilde{x}_{\cdot, j} = \frac{2 \cdot (x_{\cdot, j} - \min(x_{\cdot, j}))}{\max(x_{\cdot, j}) - \min(x_{\cdot, j})} - 1,\quad j = 1,\dots, 8.
    \end{equation*}
\end{itemize}
To fit a NN, we use the \texttt{R} package \texttt{keras}. Once the NN is trained, we extract the weights of its last hidden layer and use them as inputs to fit a Poisson GLM. The search for the optimal hyper-parameters of the NN is based on the KPIs introduced in Subsection \ref{subsec:OutperformingTheBenchmarkGLMViaRegularizedNeuralNetwork}. To find the best CANN model, we search for the best combination of hyper-parameters along a pre-defined grid of hyper-parameter. We focus on the leaky rectified linear unit (LReLU), sigmoid ($\sigma$) and hyperbolic tangent (TanH) activation functions that are defined as
\begin{align*}
    \text{LReLU}({w},\alpha) &= \max({w}, \alpha \cdot {w}),\text{ }\\
    \sigma({w}) &= \frac{1}{1+e^{-{w}}}, \text{ }\\
    \text{TanH}({w}) &= 2\sigma(2{w})-1,
\end{align*}
with ${w}$ denoting the weighted sum of the inputs of the neuron to which the activation function is applied and $\alpha \in (0, 1)$ is a parameter, which we set to $0.3$ in all our case studies. We use the Poisson deviance loss function, which is minimized via the RMSProp optimizer. To prevent overfitting, we use drop out rate of $5\%$ and early stopping of model training when the value of the loss function is not improved $5$ epochs in a row. We set the dimension of embedding layers to $2$, $q_1 = 20$, $q_2 = 15$, $q_3 = 10$. Among activation functions, we focus on LReLU, sigmoid, and TanH.  In addition, the batch size is set to $1000$ and the number of epochs to $100$.

\begin{figure}[H]
    \centering
  \includegraphics[width=\textwidth]{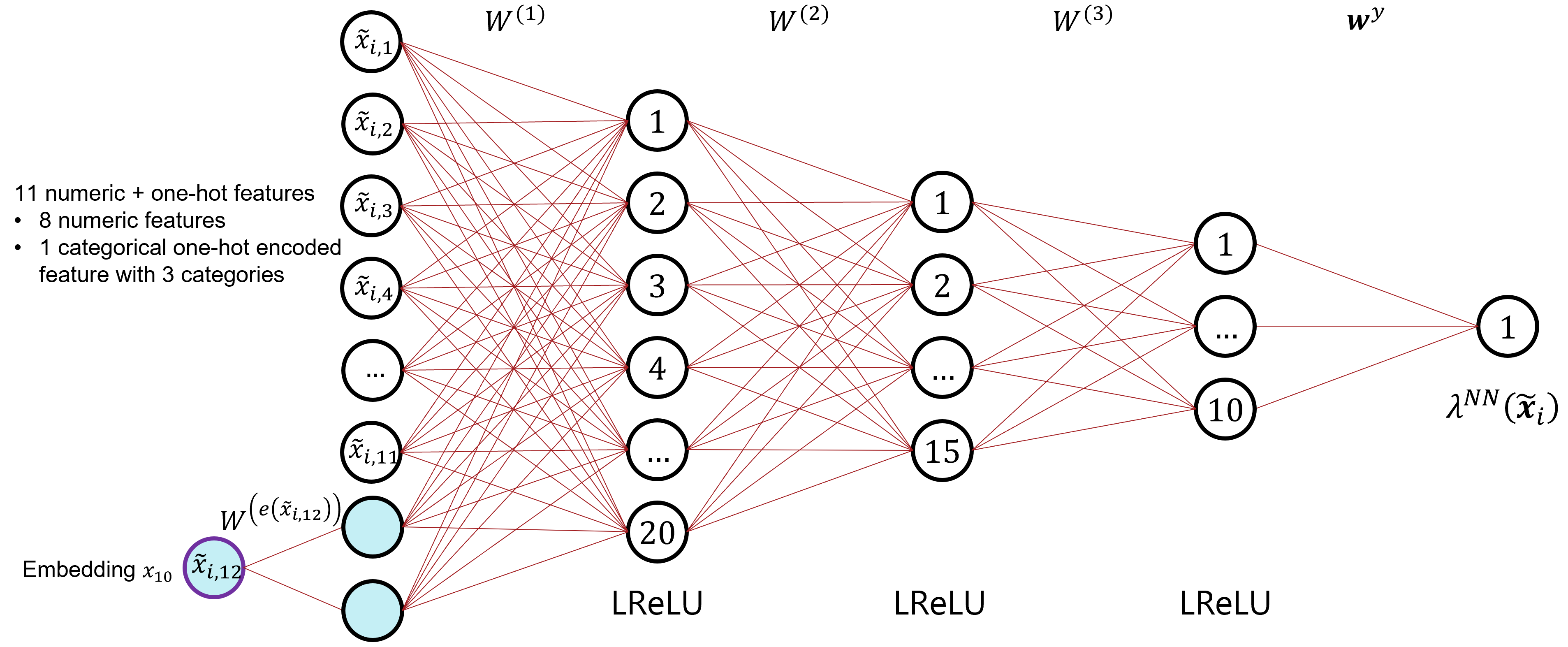}
  \caption{The architecture of the NN component of the best-performing CANN model.} 
  \label{fig:Artificial_BestNNGLMArchitecture}
\end{figure}

According to both Poisson deviance and the lift-plot-based KPIs \texttt{mae\_lift\_pb} and \texttt{mae\_lift\_qbb}, the best CANN model (among those we tested) for the artificial data has LReLU activation function in all neurons of all hidden layers. This architecture is summarized in Figure \ref{fig:Artificial_BestNNGLMArchitecture}.

As can be seen, the input layer is composed of $13$ neurons corresponding to the $8$ numeric features and the one-hot encoded feature capturing $3$ categories. Moreover, the categorical feature $x_{10}$ is encoded via an embedding layer of dimension two. The input layer is connected to the first hidden layer via the weight matrix $W^{(1)}$. Similarly, the first and the second hidden layers are connected via weight matrix $W^{(2)}$, the second and the third hidden layers are connected via weight matrix $W^{(3)}$. The third (last) hidden layer is connected to the output layer via the vector of weights $\pmb{w}^{y}$.

The KPIs for the best CANN are summarized in Table \ref{tab:Artificial_BestNNGLM_kpis}.

\begin{table}[H]
 \caption{KPIs on the test data for the best-performing CANN model}
 \label{tab:Artificial_BestNNGLM_kpis}
 \begin{tabular*}{\textwidth}{llllllll}
    \hline\noalign{\smallskip}
     Pois. dev. & Pois. dev. bench. & lift\_pb & lift\_pb\_benchm. & lift\_qbb & lift\_qbb\_benchm.\\
    \hline\noalign{\smallskip}
     0.3049 & 0.3314 & 0.0033 & 0.0277 & 0.0042 & 0.0278\\
    \noalign{\smallskip}\hline
 \end{tabular*}
\end{table}

\subsubsection{Step 2: Ranking of learned interactions via neural interaction detection}

From the best-performing CANN model, we extract the weight matrices $W^{(1)}$, $W^{(2)}$, $W^{(3)}$ as well as the vector $\pmb{w}^y$. The weight matrices can be extracted using the \texttt{get\_weights} function in \texttt{R}. The structure of the resulting output is shown in Listing \ref{lst:Artificial_WeightMatrices}. The first element of the list corresponds to the embedding weight matrix $W^{(e(\tilde{x}_{12}))} \in \mathbb{R}^{7 \times 2}$ depicted in Figure \ref{fig:Artificial_BestNNGLMArchitecture}. The second list element represents the transposed version of the weight matrix $W^{(1)}$ connecting the input layer consisting of $13$ neurons ($11$ numeric \& one-hot features + $1\cdot2$ neurons of the embedding layer related to $x_{10}$) with the first hidden layer, which has $20$ neurons. Likewise, the fourth element of the list and the sixth one correspond to the transposed versions of the weight matrices $W^{(2)}$ and $W^{(3)}$, respectively. The eighth element of the list is the vector $\pmb{w}^y$ that connects the last hidden layer of the NN component with its output neuron $y$. The third, the fifth, the seventh and the ninth element of the list each represent bias vectors corresponding to the three hidden layers and the output layer of the NN, respectively. The tenth and eleventh element of the list are the non-trainable weights of the NN component (see red connections in Figure \ref{fig:CANN_example}). The twelfth (last) element of the list is the (non-trainable) bias element related to the output neuron of CANN.

\begin{lstlisting}[caption={Weight matrices}, label={lst:Artificial_WeightMatrices}]
> model_weights<-get_weights(model)
> str(model_weights)
List of 12
 $ : num [1:7, 1:2] 0.0358 0.063 0.0683 0.0792 0.0507 ...
 $ : num [1:13, 1:20] 0.0656 0.0494 -0.9771 0.3547 -0.2274 ...
 $ : num [1:20(1d)] 0.049 -0.0885 -0.1021 0.0257 -0.2505 ...
 $ : num [1:20, 1:15] 0.287 -0.234 0.371 -0.104 0.799 ...
 $ : num [1:15(1d)] -0.0183 -0.02 -0.1593 0.0207 0.0435 ...
 $ : num [1:15, 1:10] 0.2464 -0.0183 -0.4323 0.5566 0.1632 ...
 $ : num [1:10(1d)] 0.083 -0.0328 0.0171 0.0682 0.0687 ...
 $ : num [1:10, 1] -1.165 0.674 0.579 0.641 -0.802 ...
 $ : num [1(1d)] 0.138
 $ : num [1, 1] 1
 $ : num [1, 1] 1
 $ : num [1(1d)] 0
\end{lstlisting}

Next, we apply the modified NID to calculate the strength of interactions for each pair of features. Following the recommendation of \cite{tsang2017detecting}, we use $\min(\cdot)$ as a surrogate function $\mu(\cdot)$. Having obtained the strength of interactions for each pair of input neurons, we apply the aggregation procedure for categorical features, as proposed in Subsection \ref{subsec:OpeningTheBlackBox:RankingLearnedInteractions}. In particular, we use minimum as the aggregation function\footnote{Using the arithmetic average as the aggregation function for categorical variables does not change the ranking of top $5$ interactions}. Finally, we sort the resulting list and provide top $5$ entries in Table  \ref{tab:Artificial_NIDResultsAggregated}.

\begin{table}[H]
 \caption{NID results aggregated}
 \label{tab:Artificial_NIDResultsAggregated}
 \begin{tabular*}{\textwidth}{lllr}
    \hline\noalign{\smallskip}
    Interaction rank & Feature 1 name & Feature 2 name & Interaction strength score \\
    \hline\noalign{\smallskip}
    1 & {\color{blue}$x_4$} & {\color{blue}$x_5$} & $70.0263$ \\
    2 & {\color{blue}$x_5$} & {\color{blue}$x_6$} & $37.3492$ \\
    3 & $x_4$ & $x_6$ & $34.7608$ \\
    4 & $x_5$ & $x_{10}$ & $24.3280$ \\
    5 & $x_4$ & $x_{10}$& $23.9654$ \\
    \noalign{\smallskip}\hline
 \end{tabular*}
\end{table}

As can be seen, our modified NID procedure ranks the interactions between features $x_4$ and $x_5$ and between features $x_5$ and $x_6$ as the first and the second respectively. Interestingly, the NID procedure suggests that the third-ranked interaction happens between $x_4$ and $x_6$. The reason for it is that $x_4$ appears in two interactions:  $0.5\cdot x_4 x_5 +0.125\cdot x_5^2 x_6$. The strength of interactions among other variables is quantified as much lower.

Next we compare our method with another approach used by practitioners, namely training a gradient boosting machine (GBM) and calculating Friedman's H-statistic for each pair of features. Training one GBM model takes around $120$ seconds. Calculating Friedman's H-statistic is very time-consuming for the whole data set. Therefore, we consider only a small portion of data, namely $10^4$ observations, which is $0.5\%$ share of all data. In this case, the calculation takes about $40$ seconds. We report the results in Table \ref{tab:Artificial_Hstatistic}.


\begin{table}[H]
 \caption{H-statistic results}
 \label{tab:Artificial_Hstatistic}
 \begin{tabular*}{\textwidth}{lllr}
    \hline\noalign{\smallskip}
    Interaction rank & Feature 1 name & Feature 2 name & Interaction strength score \\
    \hline\noalign{\smallskip}
    1 & {\color{blue}$x_4$} & {\color{blue}$x_5$} & $0.8495$ \\
    2 & {\color{blue}$x_5$} & {\color{blue}$x_6$} & $0.2223$ \\
    3 & $x_3$ & $x_5$ & $0.0156$ \\
    4 & $x_3$ & $x_6$ & $0.0055$ \\
    5 & $x_3$ & $x_4$& $0.0001$ \\
    \noalign{\smallskip}\hline
 \end{tabular*}
\end{table}

According to Table \ref{tab:Artificial_Hstatistic}, the true interactions have the largest H-statistic and are, thus, the strongest ones according to the method of training a GBM model and calculating Friedman H-statistic for all possible pairs of variables. However, a different amount of data may lead to a different computation time and may result in a different ranking. For example, the calculation of this interaction-strength measure for the same GBM model but using $5\%$ of data ($10^5$ observations) took about $350$ seconds and indicated a few strong but false interactions, e.g., interactions between variables $x_1$ and $x_2$, $x_7$ and $x_8$ had the H-statistic of $1$.

We would like to close this subsubsection with a brief comparison of two methods. According to \cite{molnar2019interpretable}, Friedman H-Statistic:
\begin{enumerate}
    \item can be applied to any model;
    \item is defined through the partial dependence decomposition and calculates the share of variance that is explained by the interaction;
    \item is usually (but not always) between $0$ and $1$, which allows for comparison across different models;
    \item detects all forms of interactions, independently of their specific structure;
    \item can be used for quantifying the strength of higher-order interactions, i.e., the interaction among $3$ or more features
    \item is computationally time-consuming;
    \item may lead to unstable results if not all data points are used, as the estimates also vary from run to run, which is why it is recommend to compute the H-statistic multiple times;
    \item does not provide a clear answer whether the interaction is statistically significant and it is not clear whether H-statistic is large enough to consider an interaction \enquote{strong};
    \item does not give the functional form of the interaction;
    \item has the assumption that features can be shuffled independently, which is, however, violated if features are strongly correlate strongly;
    \item may yield unexpected results for small amount of data.
\end{enumerate}

Our approach of applying NID method to the CANN model:
\begin{enumerate}
    \item is model specific and works only for feed-forward NNs with some regularity conditions on activation functions;
    \item is based on the decomposition of the strength of interaction between input neurons into two parts: the strength of connections from those input-layer neurons to the neurons in the first hidden layer, the influence of neurons in the first-hidden layer on the output neuron of the NN;
    \item does not lead to the interaction-strength score that is normalized between $0$ and $1$, which makes it challenging to compare the scores across different NNs;
    \item detects all forms of interactions learned by the NN, independently of their specific structure;
    \item can be used for quantifying the strength of higher-order interactions, i.e., the interaction among $3$ or more features;
    \item is computationally fast, since it requires only cumulative matrix multiplications of the matrices with absolute values of trained weights in the NN;
    \item always leads to the same result, given that the NN is fixed, since the method does not explicitly use data points;
    \item does not provide a clear answer whether the interaction is statistically significant and it is not clear whether the NID score is large enough to consider an interaction \enquote{strong};
    \item does not give the functional form of the interaction;
    \item has the assumption that the interactions are learned by the neural network and happen in the first hidden layer.
\end{enumerate}

\subsubsection{Step 3: Recommendation of the next-best interaction}

As described in Subsection \ref{subsec:IdentificationOfTheNextBestInteractionForAGLM}, for each interaction from Table \ref{tab:Artificial_NIDResultsAggregated} we fit a mini-GLM and keep track of the corresponding KPIs.

The mini-GLM based on the interaction between features $x_4$ and $x_5$ has the lowest AIC and the lowest residual deviance among all $5$ mini-GLMs. Therefore, it is selected as the next-best interaction to be included in the benchmark GLM.

The addition of the interaction between features $x_4$ and $x_5$ to the benchmark GLM improves the performance of the benchmark GLM. Its residual deviance drops from $596992$ to $561969$ and its AIC decreases from $804445$ to $769424$, implying that the model with interaction should be favored. The Poisson deviance on the test data drops from $0.3314$ to $0.3134$.

After the benchmark GLM has been updated by adding the recommended next-best interaction $x_4$ and $x_5$, we can repeat the whole process. Namely, training a new CANN model that uses the predictions of the updated benchmark GLM and applying the NID method to the NN component of the trained CANN model, we obtain the ranking of learned interactions as shown in Table \ref{tab:Artificial_NIDResultsAggregated_Step_2}. We see that the true interaction between features $x_5$ and $x_6$ is ranked as the strongest one. It has a much higher score than others. For each of the $5$ top-ranked interactions, we train a mini-GLM with simple parametric forms of the interaction, which are described at the end of Section \ref{sec:GLM}. As expected, the winning mini-GLM is related the true interaction between features $x_5$ and $x_6$. This model has an AIC of $769406$ and a residual deviance of $561960$ on $1800737$  degrees of freedom. The coefficient near the interaction term is significant with $p$-value of $0.0594$. The Poisson deviance on the test set is $0.3134$.

If we train mini-GLMs with a larger class of parametric forms for interactions, namely, $I(x_{\cdot,j}, x_{\cdot,k}) = \beta_{j,k} x_{\cdot,j}^{a} \cdot x_{\cdot,k}^b$ for $a \in \{ 1,2\}$ and $b \in  \{ 1,2\}$, the best-performing mini-GLM corresponds to the interaction of the form $I(x_{\cdot,5}, x_{\cdot,6}) = \beta_{5,6} \cdot x_{\cdot,5}^2 \cdot x_{\cdot,6}$.  This mini-GLM has an AIC of $763910$ and a Poisson deviance of $0.3105$ on the test set. 
Adding this interaction to the benchmark GLM leads to an AIC of $763805$ and a Poisson deviance of $0.3104$ on the test set. 

\begin{table}[H]
 \caption{NID results aggregated}
 \label{tab:Artificial_NIDResultsAggregated_Step_2}
 \begin{tabular*}{\textwidth}{lllr}
    \hline\noalign{\smallskip}
    Interaction rank & Feature 1 name & Feature 2 name & Interaction strength score \\
    \hline\noalign{\smallskip}
    1 & {\color{blue}$x_5$} & {\color{blue}$x_6$} & $25.1675$ \\
    2 & $x_3$ & $x_{10}$ & $18.8742$ \\
    3 & $x_3$ & $x_{5}$ & $17.8092$ \\
    4 & $x_3$ & $x_{6}$ & $15.6298$ \\
    5 & $x_6$ & $x_{10}$& $14.8482$ \\
    \noalign{\smallskip}\hline
 \end{tabular*}
\end{table}

To justify that our approach does not only work as desired but is additionally way more time efficient, we measure the time required for executing the above described steps of training the CANN model and applying the NID technique and the fitting of mini-GLMs. This yields on average approximately $170.3$ seconds for the training of one CANN architecture, $1.19$ seconds for the application of NID and $6.7$ seconds for fitting the mini-GLM to one interaction.


In this case study, we have verified that our methodology leads to a correct recommendation of the next-best interaction for the benchmark GLM. In the next case study, we work with a real-world open-source data set that has more features than in the toy example considered before.


\subsection{Open-source data set \texttt{freMTPL2freq}}

In this subsection, we work with an open-source data set \texttt{freMTPL2freq}, which is a part of the \texttt{R} package \texttt{CASdatasets}. We choose this data set, since it has been analysed in several papers, e.g., \cite{schelldorfer2019}, \cite{wuethrich2019}, \cite{wuethrich2020}, \cite{wuethrich2022}. We take \cite{schelldorfer2019} as main reference and use the benchmark GLM as indicated on page $5$ in \cite{schelldorfer2019}. Afterwards, we apply our interaction-detection methodology and compare our results with those stated in Section 3.5 of \cite{schelldorfer2019}.

The data set consists of $678013$ observations. Listing \ref{lst:french_data} provides a glimpse on the data.

\begin{lstlisting}[caption={Structure of the data set}, label={lst:french_data}]
> str(freMTPL2freq)
'data.frame':	678013 obs. of  12 variables:
 $ IDpol     : num  1 3 5 10 11 13 15 17 18 21 ...
 $ ClaimNb   : num  1 1 1 1 1 1 1 1 1 1 ...
 $ Exposure  : num  0.1 0.77 0.75 0.09 0.84 0.52 0.45 0.27 0.71 ...
 $ VehPower  : int  5 5 6 7 7 6 6 7 7 7 ...
 $ VehAge    : int  0 0 2 0 0 2 2 0 0 0 ...
 $ DrivAge   : int  55 55 52 46 46 38 38 33 33 41 ...
 $ BonusMalus: int  50 50 50 50 50 50 50 68 68 50 ...
 $ VehBrand  : Factor w/ 11 levels "B1","B10","B11",..: 4 4 4 4 ...
 $ VehGas    : Factor w/ 2 levels "Diesel","Regular": 2 2 1 1 1 ...
 $ Area      : Factor w/ 6 levels "A","B","C","D",..: 4 4 2 2 2 ...
 $ Density   : int  1217 1217 54 76 76 3003 3003 137 137 60 ...
 $ Region    : Factor w/ 21 levels "Alsace","Aquitaine",..: 21 21 ...
\end{lstlisting}

We conduct data pre-processing as in Section 1.3. of \cite{schelldorfer2019} and split the data into training data ($80\%$), validation data ($10\%$), and data for testing ($10\%$). Next we train the benchmark GLM, referred to as GLM2 in Section 1.3 of the mentioned paper. The resulting benchmark GLM is summarized in Listing \ref{lst:french_SummaryBenchmarkGLM}.

\begin{lstlisting}[caption={Model summary of the benchmark GLM}, label={lst:french_SummaryBenchmarkGLM}]
> summary(benchmark.GLM)

Call:
glm(formula = ClaimNb ~ VehPowerGLM + VehAgeGLM + BonusMalusGLM + 
    VehBrand + VehGas + DensityGLM + Region + AreaGLM + DrivAge + 
    log(DrivAge) + I(DrivAge^2) + I(DrivAge^3) + I(DrivAge^4), 
    family = poisson(), data = data.trainval, offset = log(Exposure))

Deviance Residuals: 
    Min       1Q   Median       3Q      Max  
-1.9626  -0.3782  -0.2891  -0.1629   6.7970  

...

    Null deviance: 200978  on 610346  degrees of freedom
Residual deviance: 190836  on 610300  degrees of freedom
AIC: 253210

Number of Fisher Scoring iterations: 6
\end{lstlisting}

\subsubsection{Step 1: Training CANN}
\label{subsubsec:Step1:OutperformingTheBenchmark}

As in the first case study, we conduct the following data pre-processing steps prior to training CANNs:
\begin{itemize}
    \item Use one-hot encoding for all categorical features with $5$ or fewer categories.
    \item Use embedding layers for all categorical features with more than $5$ categories.
    \item Apply min-max scaling to all numerical features.
\end{itemize}

We focus on CANNs with three hidden-layers such that $q_1 = 20$, $q_2 = 15$, $q_3 = 10$, and use the same grid of hyper-parameters as the one in the case study with artificially generated data.

The best-performing CANN model has LReLU activation function in all hidden layers. The KPIs of this model on the test data are summarized in Table \ref{tab:french_BestNNGLM_kpis}.

\begin{table}[H]
 \caption{KPIs on the test data for the best-performing CANN}
 \label{tab:french_BestNNGLM_kpis}
 \begin{tabular*}{\textwidth}{llllllll}
    \hline\noalign{\smallskip}
     Pois. dev. & Pois. dev. bench. & lift\_pb & lift\_pb\_bench. & lift\_qbb & lift\_qbb\_bench.\\
    \hline\noalign{\smallskip}
     $0.3636$ & $0.3749$ & $0.0067$ & $0.0326$ & $0.0096$ & $0.0317$\\
    \noalign{\smallskip}\hline
 \end{tabular*}
\end{table}

On the test data set, the best-performing CANN model outperforms the benchmark GLM in terms of all considered KPIs. This is an indication that the NN component that boosts the benchmark GLM may have found some interactions missing in the benchmark GLM. 

\subsubsection{Step 2: Ranking of learned interactions}
\label{subsubsec:Step2:RankingOfLearnedInteractions}

After training the CANN model, we apply the NID algorithm to calculate the strengths of pairwise interactions that were learned by the NN component, as described in Subsection \ref{subsec:OpeningTheBlackBox:RankingLearnedInteractions}. Similar to the case study with an artificial data set, we use minimum as a surrogate function and minimum as an aggregation function. Table \ref{tab:french_NIDResultsAggregated} summarizes the resulting strongest $10$ interactions.

\begin{table}[H]
 \caption{Top 10 interactions based on the NID}
 \label{tab:french_NIDResultsAggregated}
 \begin{tabular*}{\textwidth}{llll}
    \hline\noalign{\smallskip}
    Rank & Feature 1 & Feature 2 & NID interaction-strength score \\
    \hline\noalign{\smallskip}
      1 & VehAge & BonusMalus & $34.5956$ \\ 
      2 & VehAge & VehGas & $25.8459$ \\ 
      3 & BonusMalus & VehGas & $25.5106$ \\ 
      4 & BonusMalus & Area & $24.5642$ \\ 
      5 & BonusMalus & Region & $24.4570$ \\ 
      6  & BonusMalus & VehBrand & $23.2495$ \\
      7 & VehAge & VehBrand & $22.9104$ \\
      8 & VehAge & Region & $21.9878$ \\
      9 & DrivAge & BonusMalus & $21.1584$ \\
      10 & VehAge & Area & $20.8879$ \\
    \noalign{\smallskip}\hline
 \end{tabular*}
\end{table}

According to Table \ref{tab:french_NIDResultsAggregated}, the interaction between variables \texttt{VehAge} and \texttt{BonusMalus} is much stronger than all other pairwise interactions. The other $4$ interactions have a comparable magnitude and do not exhibit a clear \enquote{winner} among them.

Next we relate our results to those of \cite{schelldorfer2019} by reporting the interactions the researchers identified and indicating their interaction-strength rank according to our methodology: (\texttt{VehPower}; \texttt{VehAge}) with NID rank of $22$, (\texttt{VehPower}; \texttt{VehBrand}) with NID rank of $26$, (\texttt{VehAge}; \texttt{VehBrand}) with NID rank of $7$, (\texttt{VehAge}; \texttt{VehGas}) rankwith NID rank of $2$, (\texttt{DrivAge}; \texttt{BonusMalus}) with NID rank of $9$. Interestingly, the interaction between \texttt{BonusMalus} and regional variables \texttt{Area} or \texttt{Region} was not detected by the methodology proposed in \cite{schelldorfer2019}, neither was detected the interaction between \texttt{VehAge} and \texttt{BonusMalus}. 

Finally, we compare our results to the method based on GBMs and Friedman H-statistic. We choose the following grid of hyper-parameters to search for the best-performing GBM
\begin{itemize}
    \item number of trees $100$, $200$, $300$;
    \item minimal number of observations in a node $10$, $25$, $50$;
    \item shrinkage parameter $0.01$, $0.05$, $0.1$,
\end{itemize}
and train the corresponding $27$ GBM models with the benchmark-GLM prediction as an offset. Training one GBM takes on average $80$ seconds for the data under consideration. The best-performing GBM in terms of Poisson deviance has $100$ trees, $50$ as the minimal number of observations in a node, shrinkage parameter of $0.1$, and the bag-fraction parameter $0.5$. The KPIs of this model are reported in Table \ref{tab:french_BestGBM_kpis}. Interestingly, the best-performing GBM model has a better Poisson deviance than the best-performing CANN model, but its lift-plot based KPIs are worse.

\begin{table}[H]
 \caption{KPIs of the best-performing GBM model on the test data}
 \label{tab:french_BestGBM_kpis}
 \begin{tabular*}{\textwidth}{llllllll}
    \hline\noalign{\smallskip}
     Pois. dev. & Pois. dev. bench. & lift\_pb & lift\_pb\_bench. & lift\_qbb & lift\_qbb\_bench.\\
    \hline\noalign{\smallskip}
     $0.3431$ & $0.3749$ & $0.0493$ & $0.0718$ & $0.0512$ & $0.0595$\\
    \noalign{\smallskip}\hline
 \end{tabular*}
\end{table}

When the whole data set is used, the calculation of Friedman H-statistic for each pair of variables takes around $5$ minutes. We report the corresponding strongest 8 pairwise interactions in Table \ref{tab:french_GBM_Friedman_scores}. The H-statistic for each of the remaining pairwise interactions is $0$.  

\begin{table}[H]
 \caption{Top $8$ interactions according GBM + Friedman H-statistic}
 \label{tab:french_GBM_Friedman_scores}
 \begin{tabular*}{\textwidth}{lllll}
    \hline\noalign{\smallskip}
    Rank (H-stat.)& Feature 1 & Feature 2 & Friedman H-statistic & Rank (NID)\\
    \hline\noalign{\smallskip}
      1 & VehAge & VehGas & $0.3436$ & 2\\ 
      2 & DrivAge & Region & $0.2728$ & 15\\ 
      3 & DrivAge & BonusMalus & $0.1660$ & 9\\ 
      4 & BonusMalus & VehBrand & $0.0997$ & 6\\ 
      5 & VehAge & BonusMalus & $0.0729$ & 1\\ 
      6 & VehGas & Region & $1.9 \cdot 10^{-14}$ & 12\\
      7 & VehBrand & Region & $1.6 \cdot 10^{-14}$ & 18\\
      8 & VehAge & DrivAge & $6.7 \cdot 10^{-15}$ & 20\\
    \noalign{\smallskip}\hline
 \end{tabular*}
\end{table}

We see that the first-strongest pairwise interaction according to the Friedman H-statistic is the second strongest interactions according to NID. The first-strongest pairwise interaction according to our approach is ranked as the fifth strongest according to GBM \& Friedman H-statistic. Interestingly, the pairwise interactions between \texttt{BonusMalus} and regional variables is not captured by the approach of GBM \& Friedman H-statistic. 


\subsubsection{Step 3: Recommendation of the next-best interaction}
\label{subsubsec:Step3:Recommendation}

As described in Subsection \ref{subsec:IdentificationOfTheNextBestInteractionForAGLM}, for each interaction from Table \ref{tab:french_NIDResultsAggregated} we fit a mini-GLM and keep track the KPIs of interest. All mini-GLMs lead to the Poisson deviance of $0.3696$ on the test set. Based on the AIC, the winning mini-GLM achieves the lowest AIC of $279859.2$, has $22$ coefficients that are significant with $p$-values $\leq 0.01$ and corresponds to the interaction between \texttt{BonusMalus} and \texttt{Region}. This interaction is then recommended to an actuary for improving the benchmark GLM.

If an actuary prefers to use another performance measure, it may well be that another interaction is recommended as the next-best one. For example, using BIC for evaluating mini-GLMs, our methodology would suggest the interaction between \texttt{VehAge} and \texttt{VehGas}, since the corresponding mini-GLM has the lowest BIC ($279941.9$) and has all coefficients significant with $p$-values $\leq 0.01$.

In contrast to the case study with the artificial data set, we do not know the true functional form of the interaction between variables. Therefore, one may want to explore more sophisticated pairwise interaction terms, as mentioned in Section 3.3 in Remark 2. All in all, the determination of the optimal functional form of the next-best interaction is beyond the scope of this paper. The final decision is to be made by the actuaries.

\subsection{Brief discussion on proprietary data sets}
\label{subsec:InsightsFromTheAnalyisOfABigConfidentialDataSet}

Data sets of large insurance companies contain millions observations (policy snippets) with dozens of features\footnote{For example, we had a chance to work with approximately $11$ million observations and over $50$ features. The calculation of Friedman H-statistic was computationally expensive, whereas our methodology was fast.}. Some of the categorical features, e.g., postcode or vehicle model, have a high number of categories. In such cases, our methodology is especially powerful. Due to a very large number of possible pairwise interactions, comparing all of them by training as many mini-models or refitting as many times the benchmark GLM would come with huge time costs. An alternative method of finding the best-performing GBM model that uses the benchmark-GLM predictions as offset and then evaluating the strength of all interactions via Friedman H-statistic is very time-consuming, as we already saw in the cases studies for smaller open-source data sets. Our approach to interaction detection is instantaneous, once the CANN model is trained. Moreover, embedding layers in the trained CANN model allow to efficiently cluster categories of categorical variables with a large number of categories (e.g., postcodes, car brands) to be able to include them in the benchmark GLM.

\section{Conclusion}

In this paper, we proposed an approach to detecting the next-best interaction missing in an arbitrary but fixed benchmark GLM. Even though our context was MTPL insurance claim frequency, the approach can be used for other insurance business lines where GLMs or GAMs are used. The first step is training a Combined Actuarial Neural Network model, which can be seen as boosting the benchmark GLM by a neural network. The second step is ranking learned pairwise interactions by their strength, which is quantified by our modification of a fast model-specific Neural Interaction Detection approach. The third step is identifying the next-best interaction by training and comparing a small number of mini-GLMs that correspond to the top-ranked interactions. In the case studies, we validated our approach on two different data sets and compared it with the alternative approach of training a Gradient Boosing Machine with trees as weak learners and calculating Friedman H-statistic for ranking the learned interactions by their strength.

There are several advantages of our methodology. First, it is faster than other approaches based on Friedman H-statistic. Therefore, our methodology is especially suitable for big data sets with dozens of features and millions of observations. Since our approach is a fully automatable and data-driven way of enhancing a benchmark GLM, it can substantially decrease the amount of time that pricing actuaries spend on searching for interactions to improve their GLMs, which is often time-consuming. Second, by means of embedding layers, our methodology reduces the dimensionality of categorical variables with a large number of unique categories (e.g., postcode, car maker). Clustering of these embedding representations by standard clustering algorithms provides actuaries with an alternative grouping of categories of such categorical variables. This alternative can be beneficial for further improvements of benchmark GLMs.

The proposed interaction-detection methodology has several degrees of freedom, e.g., the encoding of features, hyper-parameters of the NN, the clustering algorithm and the measure for evaluation of clustering results, the KPIs for selecting the best-performing CANN, and those for comparing mini-GLMs. Therefore, it would be interesting to analyze how sensitive our  approach is to different choices for each degree of freedom.

\begin{acknowledgements}
Yevhen Havrylenko acknowledges that the major part of research presented in this paper was done during his work at the Technical University of Munich. Both authors acknowledge the support of ERGO Center of Excellence in Insurance, funded by the ERGO Group AG. We thank Kay Adam for providing the data as well as for valuable suggestions and Frank Ellgring for the opportunity to gain practical insights in actuarial pricing at Global P\&C Pricing Department at ERGO Group AG. We acknowledge the support of Noel Stein, Samarth Mehrotra, Mario Ponce-Martinez, and Yichen Lou in the preparation phase of this project. 
\end{acknowledgements}

\bibliographystyle{spmpsci}      
\bibliography{GLM_Enhancement_using_NID.bib}   


%

\section*{Conflict of interest}
The authors declare that they have no conflict of interest.

\begin{appendix}

\section{Genetic algorithms for fine-tuning neural networks}\label{app:genetic_algorithms}
Genetic algorithm (GA) is an approach to solving complex optimization problems. This algorithm belongs to the class of evolutionary algorithms and is commonly used to find high-quality (near-optimal) solutions in optimization problems. The algorithm was inspired by Darwin's idea of natural selection.

In a GA, a population of candidate solutions to an optimization problem evolves toward better candidate solutions, also called individuals. Each candidate solution has a set of properties, also called genotype. This properties can be mutated and changed; traditionally, candidate solutions are encoded using vectors of $0$ and $1$.

At the beginning of the evolution process, a population of randomly generated individuals is generated. The population in each iteration is called a generation. In each generation, the fitness of each individual in the population is evaluated, which is commonly measured as the value of the objective function in the optimization problem to be solved. Then the more fit candidate solutions are selected from the current population. Their properties (genotypes) are combined and mutated to produce a new candidate solution. This way a new generation appears, which is then used in the next iteration of the algorithm. The algorithm stops when an individual with a satisfactory fitness level is found or when a maximum number of generations has been produced.
So GA consists of three basic operations:

\begin{itemize}
    \item Selection, i.e., the determination which candidate solutions to preserve for further reproduction
    \item Crossover, i.e., the process of combining existing individuals to produce a new one based on their properties
    \item Mutation, i.e., the addition of diversity and novelty into the newly produced individual, e.g., by randomly swapping or turning-off solution bits
\end{itemize}

Let us consider an example of fine-tuning a NN, which means finding the best hyper-parameters. Assume that the GA starts with $20$ different combinations of hyper-parameters. The loss function of a NN can be used for measuring how good (fit) the candidate solution (individual) is --- the lower the loss function, the fitter the individual. The algorithm selects the best two NNs and creates an \enquote{offspring} NN that inherits the values of hyper-parameters from the  \enquote{parental} NNs. Subsequently, tiny mutations are made in the hyper-parameters of the child NN and its loss function is computed. If this loss function value is smaller than the largest loss function in the population, the child NN replaces the corresponding NN with the worst fitness. This procedure is repeated until some stopping criterion is satisfied. Finally, the GA  returns the optimal combination of hyper-parameter values from the final population of NNs.

In \texttt{R}, one can use, e.g., the package \texttt{GA} for  optimizing the hyper-parameters of NNs with the help of a genetic algorithm.

\section{Code for NID algorithm}\label{app:NID_code}
\begin{lstlisting}[caption={Code for executing NID procedure}, label={lst:NID}]
# extract NN model weight matrices
model_weights <- get_weights(CANN_model) # extract weights from CANN 
# IL = input layer, OL = output layer, HL = hidden layer
layer_1_index <- num.embed.features + 1 # from IL to HL 1
layer_2_index <- layer_1_index + 2 # from HL1 to HL2
layer_3_index <- layer_2_index + 2 # from HL2 to HL3
layer_4_index <- layer_3_index + 2 # from HL3 to OL
m1_matrix <- model_weights[layer_1_index][[1]] # transpose(W^(1))
m1_matrix_abs <- abs(m1_matrix)
m2_matrix <- model_weights[layer_2_index][[1]] # transpose(W^(2))
m2_matrix_abs <- abs(m2_matrix)
m3_matrix <- model_weights[layer_3_index][[1]] # transpose(W^(3))
m3_matrix_abs <- abs(m3_matrix)
m4_matrix <- model_weights[layer_4_index][[1]] #w^y 
m4_matrix_abs <- abs(m4_matrix)

# Compute the influence vector
influence_matrix <- m2_matrix_abs %*% m3_matrix_abs  %*% m4_matrix_abs

# Compute the interaction between the input neuron i and j
calc_interaction_strength_ij <- function(i, j, m1_matrix_abs, influence_matrix){
  # Input: index of input nodes, weight matrix (M1), influence matrix 
  # Outut: interaction between input neurons i and j
  total_interaction <- 0

  # iterate through neurons of first hidden layer, compute interaction
  # at each node and sum up interactions
  for (m in 1:length(influence_matrix)){
    w_i <- m1_matrix_abs[i,m]
    w_j <- m1_matrix_abs[j,m]
    interaction_strength <- min(w_i,w_j)
    interaction_influence <- interaction_strength * influence_matrix[m]
    total_interaction <- total_interaction + interaction_influence
  }
  return(total_interaction)
}
\end{lstlisting}

In \texttt{keras} the matrices extracted by the function \texttt{get\_weights} are the transposed versions of the matrices used in our notation as well as the notation by Tsang et al. (2018). Therefore, in the code above \texttt{influence\_matrix} corresponds to $\rBrackets{\pmb{\zeta}^{(1)}}^\top = |W^{(2)}|^\top \cdot|W^{(3)}|^\top\cdot |\pmb{w}^{y}|$ in the notation of our paper, $d = 3$.

\end{appendix}

\end{document}